%% file: main.tex
\documentclass[10pt,twocolumn,letterpaper]{article}

\usepackage[pagenumbers]{cvpr} 

\input{preamble}

\definecolor{cvprblue}{rgb}{0.21,0.49,0.74}
\usepackage[pagebackref,breaklinks,colorlinks,allcolors=cvprblue]{hyperref}

\def\paperID{10423} 
\def\confName{CVPR}
\def\confYear{2026}

\title{Collaborative Learning with Multiple Foundation Models for \\Source-Free Domain Adaptation}

\author{Huisoo Lee$^{1}$ \quad Jisu Han$^{2}$ \quad Hyunsouk Cho$^{1}$ \quad Wonjun Hwang$^{2}$\\
$^{1}$Ajou University \quad $^{2}$Korea University\\
{\tt\small huisu0818@ajou.ac.kr jisuhan@korea.ac.kr hyunsouk@ajou.ac.kr wjhwang@korea.ac.kr}
}

\begin{document}
\maketitle
\input{sec/0_abstract}    
\input{sec/1_intro}
\input{sec/2_related_work}
\input{sec/3_methodology}
\input{sec/4_experiments}
\input{sec/5_conclusion}
{
    \small
    \bibliographystyle{ieeenat_fullname}
    \bibliography{main}
}

\input{sec/6_supplementary}

\end{document}

%% file: preamble.tex
\definecolor{best}{RGB}{208, 103, 130}
\newcommand{\best}[1]{{\textbf{#1}}} 

\definecolor{bestrow}{RGB}{220, 232, 248} 

\usepackage[table]{xcolor}
\usepackage{multirow}
\usepackage{pifont} 
\usepackage{algorithmic,algorithm}
\newcommand{\cmark}{\ding{51}}
\newcommand{\xmark}{\ding{55}}

\newtheorem{definition}{Definition}
\newtheorem{proposition}{Proposition}

%% file: sec/0_abstract.tex
\begin{abstract}
Source-Free Domain Adaptation (SFDA) aims to adapt a pre-trained source model to an unlabeled target domain without access to source data.
Recent advances in Foundation Models (FMs) have introduced new opportunities for leveraging external semantic knowledge to guide SFDA.
However, relying on a single FM is often insufficient, as it tends to bias adaptation toward a restricted semantic coverage, failing to capture diverse contextual cues under domain shift.
To overcome this limitation, we propose a \textbf{Co}llaborative \textbf{M}ulti-foundation \textbf{A}daptation (\textbf{CoMA}) framework that jointly leverages two different FMs (e.g., CLIP and BLIP) with complementary properties to capture both global semantics and local contextual cues.
Specifically, we employ a bidirectional adaptation mechanism that (1) aligns different FMs with the target model for task adaptation while maintaining their semantic distinctiveness, and (2) transfers complementary knowledge from the FMs to the target model.
To ensure stable adaptation under mini-batch training, we introduce Decomposed Mutual Information (DMI) that selectively enhances true dependencies while suppressing false dependencies arising from incomplete class coverage.
Extensive experiments demonstrate that our method consistently outperforms existing state-of-the-art SFDA methods across four benchmarks, including Office-31, Office-Home, DomainNet-126, and VisDA, under the closed-set setting, while also achieving best results on partial-set and open-set variants.
\end{abstract}

%% file: sec/1_intro.tex
\section{Introduction}
\label{sec:intro}

\begin{figure}[t] 
    \setlength{\belowcaptionskip}{-10pt}
    \setlength{\abovecaptionskip}{-2pt}
    \begin{center}
     \includegraphics[width=0.95\linewidth]{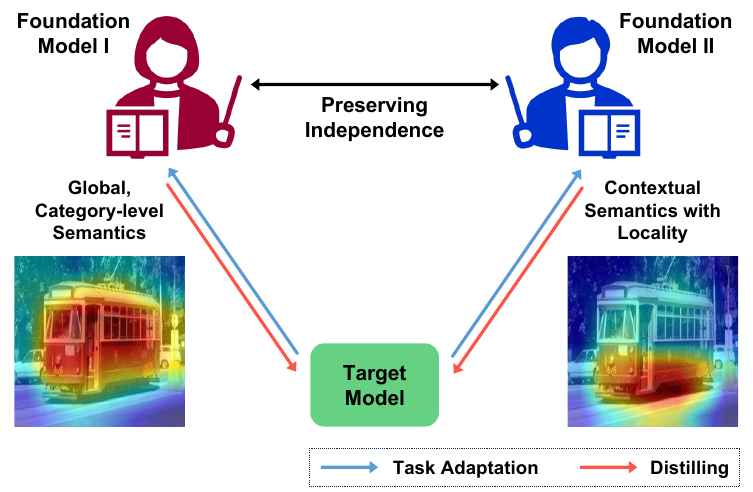}
    \end{center}
     \caption{
    While prior SFDA methods typically employ a single FM, we jointly leverage two different FMs with complementary semantic properties. The visualization illustrates how FM I (e.g., CLIP) captures global, category-level semantics (e.g., ``tram"), whereas FM II (e.g., BLIP) captures local contextual semantics with fine-grained details (e.g., ``wheels")~\cite{clip,instblip,clip_semantic1,clip_semantic2}. The target model guides FMs toward task-relevant semantics and then obtains their refined signals for improving its prediction.
    }
    \label{fig:idea-comp}
\end{figure}

Unsupervised Domain Adaptation (UDA) aims to transfer knowledge from a labeled source domain to an unlabeled target domain. However, accessing to source data is often infeasible in real-world scenarios due to privacy or legal constraints~\cite{SFDA}.
Source-Free Domain Adaptation (SFDA) addresses this issue by adapting a pretrained source model to the target domain using only unlabeled target data, without accessing the source data~\cite{shot}.
Due to the absence of source data and target labels, most prior methods mainly rely on self-supervised objectives such as entropy minimization~\cite{shot,shot++,Uncertainty,tpds}, contrastive learning~\cite{aanet,apg,adacon,plue,crs} or neighborhood clustering~\cite{gsfda,nrc,aad} to refine the target predictions.
While these strategies can gradually align target features, their effectiveness is limited by the uncertainty of the source model under domain shift.

Recent studies~\cite{difo,prode} have introduced Multimodal Foundation Models (MFMs) as external semantic guidance to compensate for the limited generalization of the source model and reduce reliance on noisy pseudo-labels.
However, their ability to exploit local visual cues remains limited, particularly for MFMs such as CLIP~\cite{clip}.
Trained on large-scale image–text pairs, CLIP captures broad visual–language associations that support strong zero-shot transfer but lack sensitivity to fine-grained details~\cite{clip_semantic1}.
As a result, its representations often overlook local information such as boundaries, spatial relations, or textures, leading to misaligned or incomplete semantics under domain shifts.

In this paper, we propose a \textbf{Co}llaborative \textbf{M}ulti-foundation \textbf{A}daptation (\textbf{CoMA}) framework that jointly leverages two complementary MFMs (e.g., CLIP and BLIP~\cite{instblip}).
As shown in Fig.~\ref{fig:idea-comp}, we establish both global and local semantic grounding, bridging them with the target model for the SFDA task.
It first aligns two MFMs toward task-relevant semantics while preserving their semantic distinctiveness under the guidance of the target model, then transfers complementary knowledge from MFMs to the target model through agreement-guided supervision and selective information maximization.
This bidirectional learning mechanism achieves both cross-model consistency and intra-model reliability without collapsing the diverse semantic signals provided by two MFMs.
To effectively realize these objectives, we introduce Decomposed Mutual Information (DMI), a selective information-theoretic formulation that addresses the instability of conventional Mutual Information (MI) under mini-batch training by selectively enhancing true dependencies while suppressing false ones.

Our \textbf{contributions} are summarized as follows:
\begin{itemize}
    \setlength{\leftskip}{12pt}
    \item To the best of our knowledge, we are the first to jointly leverage multiple MFMs with complementary semantic properties in a bidirectional and cross-model learning framework for the SFDA problem.
    \item We propose DMI, a novel information-theoretic formulation that decomposes the joint distribution into confident and uncertain regions, selectively enhancing true dependencies while 
    suppressing false dependencies for stable knowledge transfer.
    \item We design a CoMA framework that bridges different MFMs with the target model through bidirectional interaction, achieving consistent cross-model alignment and reliable target adaptation.
\end{itemize}

%% file: sec/2_related_work.tex
\section{Related Work}
\label{sec:related_work}

\begin{figure*}[t]
    \setlength{\belowcaptionskip}{-5pt}
    \setlength{\abovecaptionskip}{-1pt}
    \begin{center}
        \includegraphics[width=0.95\linewidth,angle=0]{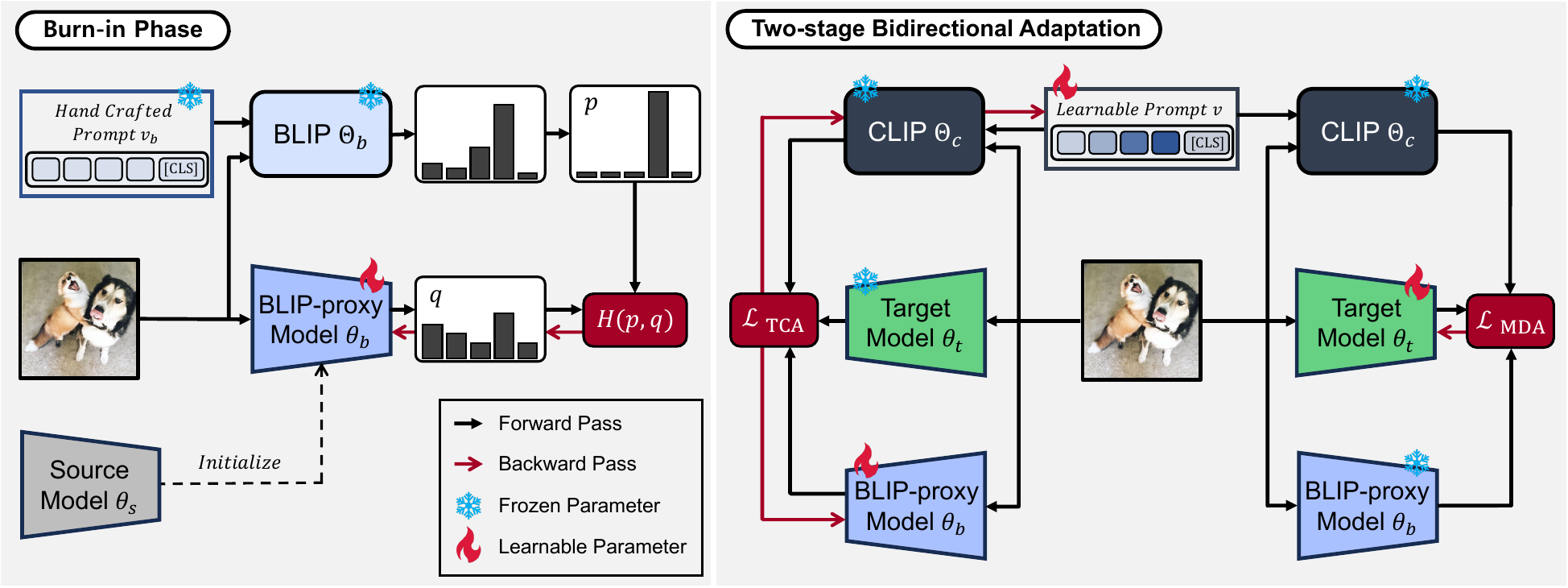}
    \end{center}
    \caption{
    Overview of \textbf{CoMA}. Our method begins with a burn-in phase that trains a BLIP-proxy model initialized from the source model.
    After this phase the target model is also initialized from the source model. It then undergoes two bidirectional stages:
    (1) TCA aligns two complementary MFMs with task-relevant semantics while preserving their semantic distinctness, and
    (2) MDA transfers reliable knowledge from MFMs into the target model.
    $H(p,q)$ denotes cross-entropy.
    }
    \label{fig:fw}
\end{figure*}

\textbf{Source-free domain adaptation.}
Existing SFDA approaches can be broadly categorized into two major streams: data-based and model-based methods~\cite{survey}. Recently, a new line of research has emerged that leverages external information as an additional knowledge source.

Data-based methods exploit the intrinsic structure of unlabeled target data to induce discriminative and domain-invariant feature representations. For example, neighborhood clustering approaches~\cite{gsfda,nrc,aad} utilize local consistency among target samples to form domain-invariant neighborhoods, improving feature discrimination within the target space. Meanwhile, virtual domain estimation methods~\cite{vdm,cowa,proxyMix,gaa} reconstruct or approximate the source domain distribution by generating pseudo-source representations, thereby facilitating target alignment without direct source data access.

Model-based methods perform self-supervised adaptation by optimizing prediction consistency and information-level objectives using the pretrained source model itself. Specifically, entropy minimization approaches~\cite{shot,shot++,Uncertainty,tpds} refine target representations by promoting confident and diverse predictions.
Contrastive learning methods~\cite{aanet,apg,adacon,plue,crs} enhance inter-sample dependency modeling by maximizing mutual information across feature instances, encouraging compact intra-class and dispersed inter-class structures for more discriminative alignment.

Despite these advancements, conventional SFDA methods remain fundamentally limited by the inevitable errors derived from the source model and unlabeled target data under distribution shift~\cite{difo}.
To overcome this constraint, recent works~\cite{difo,prode} incorporate an MFM to provide external semantic guidance. Trained on large-scale multimodal data, MFMs offer more robust and generalizable knowledge beyond the scope of domain-specific source models.

\vspace{4pt}
\noindent\textbf{Multimodal foundation model.}
MFMs such as CLIP, ALIGN \cite{ALIGN} and GLIP \cite{glip} have achieved remarkable success by aligning visual and textual modalities through large-scale contrastive pretraining. These models learn modality-invariant representations that generalize across diverse downstream tasks, making them a promising external knowledge source for domain adaptation.

It is worth noting that different MFMs encode semantics at distinct levels of granularity.
CLIP captures global vision–language associations through image–text contrastive learning, offering strong category-level alignment but limited local grounding~\cite{clip_semantic1,clip_semantic2}. On the other hand, BLIP employs cross-attention mechanisms to establish detailed region–text correspondences, emphasizing local contextual semantics.

Recent SFDA methods~\cite{difo,prode} focus on only a single MFM, which restricts semantic diversity and limits robustness when the model’s intrinsic semantic granularity mismatches the target domain. Consequently, such semantic bias often results in incomplete guidance under domain shift.
To address this limitation, we jointly adopt CLIP and BLIP, which capture global and local semantics, respectively, enabling balanced semantic coverage at different levels of granularity.

%% file: sec/3_methodology.tex
\section{Methodology}

\textbf{Problem definition.}
Following the standard SFDA setting~\cite{shot}, we assume a pre-trained source model is given, while source data is unavailable during adaptation. We consider a classification task where the source and target domain share the same label space $\mathcal{C} = \{1,2,\ldots,K\}$, where $K$ represents the total number of classes. We denote source samples and labels as 
$\mathcal{X}_s=\{x_s^i\}_{i=1}^{n_s}$ and $\mathcal{Y}_s=\{y_s^i\}_{i=1}^{n_s}$, and unlabeled target 
samples and their unknown labels as $\mathcal{X}_t=\{x_t^i\}_{i=1}^{n_t}$ and $\mathcal{Y}_t=\{y_t^i\}_{i=1}^{n_t}$, where $n_s$ and $n_t$ denote the number of samples in each domain.
Our goal is to train a target model $\theta_t : \mathcal{X}_t \rightarrow \mathcal{Y}_t$ that can accurately predict the target labels $\mathcal{Y}_t$ using only the unlabeled target data $\mathcal{X}_t$ and the pretrained source model $\theta_s : \mathcal{X}_s \rightarrow \mathcal{Y}_s$. 
In addition, we utilize CLIP $\Theta_c$, BLIP $\Theta_b$, and a BLIP-proxy model $\theta_b$ 
trained on BLIP-generated pseudo-labels.

\vspace{4pt}
\noindent\textbf{Overview.}
Our framework consists of a burn-in phase followed by a two-stage bidirectional adaptation procedure as illustrated in Fig.~\ref{fig:fw}.

The burn-in phase trains a BLIP-guided proxy model $\theta_b$, which is initialized from the source model and trained with BLIP-generated pseudo-labels. It serves as an efficient surrogate for the BLIP $\Theta_B$. This design allows our framework to leverage BLIP’s semantic signals in a computationally efficient manner rather than using the full model during adaptation. Moreover, since BLIP is originally designed for caption generation and lacks a classifier head, its output space is not inherently aligned with the class-discriminative probability space used by the target model or CLIP. The proxy model bridges this gap by transforming BLIP's semantic information into classifier-compatible representations, enabling stable interaction with the target model. The implementation detail of the burn-in stage is provided in \texttt{Supplementary}~\ref{supp:burn-in}.

After burn-in, our framework proceeds in two stages within a bidirectional design:
(1) Target-Guided Consistency Adjustment (TCA), which aligns different MFMs with the target model while preserving their semantic independence, and
(2) MFM Distilled Target Adaptation (MDA), which transfers complementary knowledge from MFMs into the target model. 

To ensure stable knowledge transfer across both stages, we introduce DMI. By selectively enhancing true dependencies within confident predictions while suppressing false dependencies arising from incomplete mini-batch sampling, DMI enables robust alignment under mini-batch training.

\subsection{Decomposed Mutual Information}
\label{sec:DMI}

Mutual Information (MI) measures the dependency between two random variables $X$ and $Y$.
Formally, it can be expressed as:
\begin{equation}
\label{eq:mi}
    I(X;Y) = H(Y) - H(Y|X),
\end{equation}
where $H(\cdot)$ denotes the entropy.
As shown in Eq.~\eqref{eq:mi}, maximizing MI reduces conditional 
uncertainty $H(Y|X)$ while maintaining diversity $H(Y)$, which 
strengthens the diagonal of the joint distribution and reinforces 
consistency between distributions. Conversely, minimizing MI promotes their decorrelation, suppressing redundant dependencies.

Traditional objectives such as Kullback–Leibler (KL) divergence assume one distribution is the ground truth and directly align the other to it.
In contrast, MI does not require such a supervised assumption~\cite{difo}.
Therefore, it remains effective in unsupervised settings and has been widely adopted in unsupervised learning tasks~\cite{iic,shot,difo,prode}.

Despite its theoretical advantages, MI suffers from instability in mini-batch training due to inaccurate estimation of the joint distribution. 
As the number of classes $K$ increases, the joint matrix grows with complexity $\mathcal{O}(K^2)$, or even $\mathcal{O}(K^3)$ when conditioning on a third variable for conditional MI.
Since GPU memory constraints prevent sufficiently large batch sizes, many classes remain unobserved within each mini-batch, making the batch-level joint distribution sparse and imbalanced, particularly in datasets with large $K$.
Consequently, conventional MI maximization can inadvertently reinforce false dependencies for classes that are unobserved in the current batch. This leads to unintended alignment that degrades adaptation performance.

\begin{figure}[t] 
    \setlength{\belowcaptionskip}{-5pt}
    \setlength{\abovecaptionskip}{-2pt}
    \begin{center}
     \includegraphics[width=0.95\linewidth]{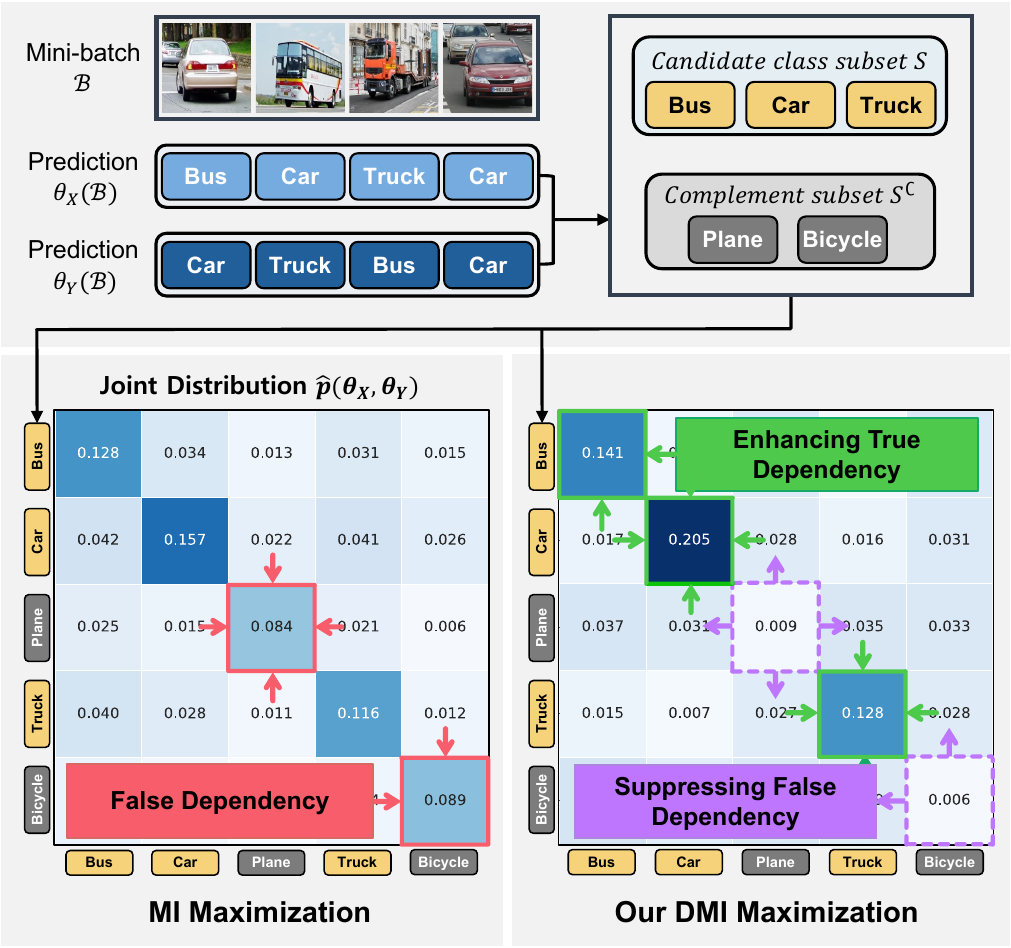}
    \end{center}
     \caption{
    Concept of our DMI method. Conventional MI maximization reinforces dependencies for all class pairs including those absent ($\mathcal{S}^{\complement}$: Plane, Bicycle) from the batch, causing false dependencies. In contrast, our DMI maximization enhances true dependencies within the confident joint region formed by classes present ($\mathcal{S}$: Bus, Car, Truck) in the batch, while suppressing false ones in the uncertain region with absent classes $\mathcal{S}^{\complement}$.
    }
    \label{fig:dmi}
\end{figure}

To address this limitation, we propose DMI, which decomposes 
the joint probability space into confident and uncertain regions. 
As illustrated in Fig.~\ref{fig:dmi}, DMI maximizes MI in the 
confident region to enhance true dependencies, while minimizing 
MI in the uncertain region to suppress false ones.

\vspace{4pt}
\noindent\textbf{Confident joint subset.}
Following the MI formulation used in representation-level objectives such as~\cite{iic},
we consider two predictive distributions produced by different models over the same mini-batch.
Let $X = \theta_X(\mathcal{B})$ and $Y = \theta_Y(\mathcal{B})$ denote the batch-wise output predictions
from two networks $\theta_X$ and $\theta_Y$, respectively, where $\mathcal{B}=\{x_t^i\}_{i=1}^{n_B}$ represents
a mini-batch sampled from $\mathcal{X}_t$ with batch size $n_B$. To stabilize MI estimation under mini-batch sampling, we selectively focus on the region of the joint space where both predictive distributions exhibit confident responses.
Such regions termed Confident Joint Subset (CJS) statistically correspond to low-entropy and high-density areas, yielding sufficient joint statistics.
In practice, we identify CJS by constructing a candidate class subset $\mathcal{S} \subseteq \mathcal{C}$ from confident predictions of both distributions within a mini-batch of target samples.
For each batch, we define:
\begin{equation}
    \begin{split}
    \label{eq:ccs}
    \mathcal{S}_X &= \{~c \mid c = \arg\max_k \theta_X(x_t),\; \forall x_t \in \mathcal{B} ~\},\\
    \mathcal{S}_Y &= \{~ c \mid c = \arg\max_k \theta_Y(x_t),\; \forall x_t \in \mathcal{B} ~\},\\
    \end{split}  
\end{equation}
where $k \in \mathcal{C}$ denotes the class index, and we obtain the candidate class subset $\mathcal{S} = \mathcal{S}_X \cup \mathcal{S}_Y$. We then denote $X_\mathcal{S}$ and $Y_\mathcal{S}$ as the subsets of $X$ and $Y$ corresponding to the class indices contained in $\mathcal{S}$. Finally, the joint space can be decomposed into $X_\mathcal{S} \times Y_\mathcal{S}$ and $X_{\mathcal{S}^{\complement}} \times Y_{\mathcal{S}^{\complement}}$, where $\mathcal{S}^{\complement}$ denotes the complement of $\mathcal{S}$ in the class space $\mathcal{C}$.
Among these, $X_\mathcal{S} \times Y_\mathcal{S}$ constitutes CJS,
where the joint signals are dense and statistically stable,
providing a noise-reduced space for learning true dependencies between the two predictive distributions.
Conversely, $X_{\mathcal{S}^{\complement}} \times Y_{\mathcal{S}^{\complement}}$ represents an unstable region containing false dependencies that can distort the estimation, and thus its MI is suppressed.

\begin{definition}
\label{def:dmi}
The Decomposed Mutual Information is formally defined as:
\begin{equation}
    I_D(X;Y) = I(X_\mathcal{S};Y_\mathcal{S}) - \frac{\log|\mathcal{S}|}{\log|\mathcal{S}^{\complement}|} \cdot I(X_{\mathcal{S}^{\complement}};Y_{\mathcal{S}^{\complement}}),
\label{eq:DMI}
\end{equation}
where $|\mathcal{S}| \ge 2$ and $|\mathcal{S}^{\complement}| \ge 2$.
\end{definition}

This formulation selectively enhances true dependencies within 
the confident region while suppressing false dependencies in the 
uncertain region, thereby stabilizing adaptation under mini-batch 
training.

\begin{proposition}
\label{propose:DMI}
For any candidate class subset $\mathcal{S}$ satisfying $|\mathcal{S}| \ge 2$ and $|\mathcal{S}^{\complement}| \ge 2$, the Decomposed Mutual Information $I_D(X;Y)$ satisfies the following bounded condition:
\begin{equation}
    -\log|\mathcal{S}| \le I_D(X;Y) \le \log|\mathcal{S}|.
\end{equation}
\end{proposition}

This proposition guarantees that $I_D(X;Y)$ remains bounded by 
$\log|\mathcal{S}|$. The scaling factor $\frac{\log|\mathcal{S}|}{\log|\mathcal{S}^{\complement}|}$ balances the enhancement and suppression terms, preventing either from dominating and yielding this symmetric bound. This ensures consistent optimization dynamics. As batch composition varies, 
the bound adapts proportionally to the size of $\mathcal{S}$, maintaining a stable optimization scale throughout training.
The proof is provided in \texttt{Supplementary}~\ref{supp:proof}.

Although CJS is constructed based on confident predictions within a mini-batch to stabilize joint estimation, the DMI formulation is not limited to this specific construction.
In principle, CJS can also be explicitly defined or reconstructed using auxiliary confidence sources, allowing the model to emphasize dependencies that align with specific training goals.
For example, in our MDA objective in Eq.~\eqref{eq:sim}, CJS is externally induced from MFM predictions rather than derived from internal target model predictions, illustrating how DMI can be tailored to different purposes.

\subsection{Target-Guided MFM Consistency Adjustment}

Although MFMs are trained for general-purpose multimodal understanding, they remain inherently task-agnostic. Previous MFM-based approaches~\cite{difo, prode} have attempted to customize a single MFM toward target-specific semantics via prompt-based adaptation~\cite{coop}. However, our setting requires a more careful approach, as it involves leveraging multiple MFMs with distinct characteristics whose semantic behaviors often diverge under domain shift. This semantic distinctiveness necessitates a unified adjustment mechanism that harmonizes both MFMs with the target model rather than customizing each individually.

To address this, we introduce the TCA stage, where the target model acts as an adaptive bridge between complementary MFMs to achieve task-consistent alignment while preserving their semantic independence.

\vspace{4pt}
\noindent\textbf{Mutual Consistency.}
To mitigate the task-agnostic nature of MFMs, we introduce the mutual consistency objective, which aligns both MFMs with the target prediction and encourages the MFMs to gradually shift from domain-general semantics toward the task-relevant semantics.
We achieve this by maximizing DMI between the target prediction and each MFM prediction:
\begin{equation}
    \label{eq:mc}
    \begin{split}
    \mathcal{L}_{\mathrm{MC}}=\min_{v,\theta_b}
   \mathbb{E}_{x_t\in \mathcal{X}_t} \big[ &-I_D\big(\theta_t(x_t);\theta_b(x_t)\big) \\
   &-\,I_D\big(\theta_t(x_t);\Theta_c(x_t,v)\big)\big],
    \end{split}
\end{equation}
where $v$ denotes learnable prompt context following~\cite{coop}.

\vspace{4pt}
\noindent\textbf{Conditional Decorrelation.}
However, simply enforcing mutual consistency may cause the two MFMs to converge toward similar predictions, losing their complementary semantic perspectives.
To mitigate this issue, we minimize conditional DMI, which measures the residual dependency between the two MFMs after accounting for the target prediction. Following the standard formulation of conditional MI, conditional DMI is computed as:
\begin{equation}
    \begin{split}
I_D(X;Y \mid Z) = \mathbb{E}_{Z}\big[ I_D(X;Y) \mid Z \big],
    \end{split}
\end{equation}
where $Z$ denotes the conditioning variable. Formally, the conditional decorrelation loss is defined as:
\begin{equation}
    \label{eq:cd}
    \begin{split}
    \mathcal{L}_{\mathrm{CD}}&=\min_{v,\theta_b}
   \mathbb{E}_{x_t \in\mathcal{X}_t}\big[
        I_D\big(\theta_b(x_t);\Theta_c(x_t,v)\mid \theta_t(x_t)\big)
    \big],
    \end{split}
\end{equation}
which prevents collapse into overly redundant predictions, helping maintain different semantics so that MFMs can provide distinct yet task-aligned information.

The entire loss for TCA stage combines both components as follows:
\begin{equation}
\label{eq:tca}
\mathcal{L}_{\mathrm{TCA}}
=\mathcal{L}_{\mathrm{MC}}
+\alpha\mathcal{L}_{\mathrm{CD}},
\end{equation}
where $\alpha>0$ balances mutual consistency and conditional decorrelation.

\subsection{MFM Distilled Target Adaptation}
\label{sec:mda}
After TCA, the two MFMs are aligned toward task-relevant semantics, while still preserving complementary perspectives.
However, the target model has not yet absorbed this refined knowledge. Thus, we introduce a distillation stage, termed MDA, to effectively transfer the complementary knowledge of MFMs into the target model.

\vspace{4pt}
\noindent\textbf{Agreement-Guided Supervision.}
The first objective exploits agreement between MFM predictions. 
When $\Theta_c$ and $\theta_b$ predict the same class, the agreed class is treated as a pseudo-label for supervising the target model. 
Since the two MFMs capture complementary semantics, their agreement indicates consensus from different semantic perspectives, ensuring confident supervision for $\theta_t$:
\begin{equation}
    \label{eq:ags}
    \mathcal{L}_{\mathrm{AGS}}
    = -\underset{\theta_t}{\min}\;
    \mathbb{E}_{(x_t, \hat{y}_t) \in \mathcal{X}_t \times \hat{\mathcal{Y}}_t}
    \sum_{k=1}^{K} p_k \log \theta_t(x_t),
\end{equation}
where $\hat{y}_t$ denotes the pseudo-label derived from MFM agreement and $p_k$ represents its one-hot encoded label.

\vspace{4pt}
\noindent\textbf{Selective Information Maximization.}
While $\mathcal{L}_{\mathrm{AGS}}$ leverages confident samples on which both MFMs agree, uncertain predictions where their outputs diverge remain unexploited. 
To address this, we guide such uncertain predictions to align according to the target model’s own semantic tendency. 
To achieve this alignment, we first construct the candidate class subset $\mathcal{S}$ explicitly, following Eq.~\eqref{eq:ccs} as the union of confident class predictions from $\Theta_c$ and $\theta_b$. 
Within this subset, the target model is encouraged to reinforce its self-consistency by maximizing MI in the manner of \cite{shot}.
Unlike the global MI optimization in~\cite{shot}, however, we reformulate it through our DMI, applied over the externally defined subset $\mathcal{S}$.
Formally, the objective is expressed as:
\begin{equation}
\label{eq:sim}
\mathcal{L}_{\mathrm{SIM}}
= -\,\mathbb{E}_{x_t \in \mathcal{X}_t}\, I_D(x_t;\,\theta_t(x_t)),
\end{equation}
which encourages the target model to produce discriminative and 
diverse predictions within the semantic scope of $\mathcal{S}$. Thus, we achieve structured MFM-guided adaptation and further align predictions, where MFMs disagree, according to the target model's own semantic tendency.

The overall MDA loss integrates the two objectives as follows:
\begin{equation}
    \label{eq:mda}
    \mathcal{L}_{\mathrm{MDA}}
    = \mathcal{L}_{\mathrm{AGS}} 
    + \beta\mathcal{L}_{\mathrm{SIM}},
\end{equation}
where $\beta>0$ is balancing coefficient.

%% file: sec/4_experiments.tex
\section{Experiments}
\begin{table*}[t]
    \caption{{\bf Closed-set SFDA accuracy} (\%) on {Office-Home}. \textbf{SF}, \textbf{C} and \textbf{B} indicate source-free, CLIP and BLIP, respectively. The best accuracy is indicated in \textbf{bold} and the second best one is \underline{underlined}.}
    \label{tab:officehome}
    \vspace{-0.15cm}
    \renewcommand\tabcolsep{1.8pt}
    \renewcommand\arraystretch{0.9}
    \scriptsize
    \centering
    \resizebox{\textwidth}{!}{
        \begin{tabular}{ l l | c c c | c c c c c c c c c c c c | c}
        \toprule
        Method & Venue & \textbf{SF} & \textbf{C} & \textbf{B} & Ar$\to$Cl & Ar$\to$Pr & Ar$\to$Rw & Cl$\to$Ar & Cl$\to$Pr & Cl$\to$Rw & Pr$\to$Ar & Pr$\to$Cl & Pr$\to$Rw & Rw$\to$Ar & Rw$\to$Cl & Rw$\to$Pr & Avg. \\
        \midrule
        Source &-- &-- &-- &-- & 43.3 & 66.1 & 73.2 & 51.5 & 60.4 & 64.0 & 53.3 & 40.4 & 73.1 & 65.3 & 45.8 & 78.4 & 59.6 \\
        \midrule
        SHOT~\cite{shot} & ICML20 & \cmark & \xmark & \xmark & 57.1 & 78.1 & 81.5 & 68.0 & 78.2 & 78.1 & 67.4 & 54.9 & 82.2 & 73.3 & 58.8 & 84.3 & 71.8 \\
        NRC~\cite{nrc} & NIPS21 & \cmark & \xmark & \xmark & 57.7 & 80.3 & 82.0 & 68.1 & 79.8 & 78.6 & 65.3 & 56.4 & 83.0 & 71.0 & 58.6 & 85.6 & 72.2 \\
        GKD~\cite{gkd}& IROS21 & \cmark & \xmark & \xmark & 56.5 & 78.2 & 81.8 & 68.7 & 78.9 & 79.1 & 67.6 & 54.8 & 82.6 & 74.4 & 58.5 & 84.8 & 72.2 \\
        AaD~\cite{aad} & NIPS22 & \cmark & \xmark & \xmark & 59.3 & 79.3 & 82.1 & 68.9 & 79.8 & 79.5 & 67.2 & 57.4 & 83.1 & 72.1 & 58.5 & 85.4 & 72.7 \\
        AdaCon~\cite{adacon} & CVPR22 & \cmark & \xmark & \xmark & 47.2 & 75.1 & 75.5 & 60.7 & 73.3 & 73.2 & 60.2 & 45.2 & 76.6 & 65.6 & 48.3 & 79.1 & 65.0 \\
        CoWA~\cite{cowa} & ICML22 & \cmark & \xmark & \xmark & 56.9 & 78.4 & 81.0 & 69.1 & 80.0 & 79.9 & 67.7 & 57.2 & 82.4 & 72.8 & 60.5 & 84.5 & 72.5 \\
        ELR~\cite{elr} & ICLR23 & \cmark & \xmark & \xmark & 58.4 & 78.7 & 81.5 & 69.2 & 79.5 & 79.3 & 66.3 & 58.0 & 82.6 & 73.4 & 59.8 & 85.1 & 72.6 \\
        PLUE~\cite{plue}& CVPR23 & \cmark & \xmark & \xmark & 49.1 & 73.5 & 78.2 & 62.9 & 73.5 & 74.5 & 62.2 & 48.3 & 78.6 & 68.6 & 51.8 & 81.5 & 66.9 \\
        CPD~\cite{cpd}& PR24 & \cmark & \xmark & \xmark & 59.1 & 79.0 & 82.4 & 68.5 & 79.7 & 79.5 & 67.9 & 57.9 & 82.8 & 73.8 & 61.2 & 84.6 & 73.0 \\
        TPDS~\cite{cpd} & IJCV24 & \cmark & \xmark & \xmark & 59.3 & 80.3 & 82.1 & 70.6 & 79.4 & 80.9 & 69.8 & 56.8 & 82.1 & 74.5 & 61.2 & 85.3 & 73.5 \\
        \midrule
        DAPL-R~\cite{daplr} & TNNLS23 & \xmark & \cmark & \xmark & 54.1 & 84.3 & 84.8 & 74.4 & 83.7 & 85.0 & 74.5 & 54.6 & 84.8 & 75.2 & 54.7 & 83.8 & 74.5 \\
        PADCLIP-R~\cite{padclipr} & ICCV23 & \xmark & \cmark & \xmark & 57.5 & 84.0 & 83.8 & 77.8 & 85.5 & 84.7 & 76.3 & 59.2 & 85.4 & 78.1 & 60.2 & 86.7 & 76.6 \\
        ADCLIP-R~\cite{adclipr}& ICCVW23 & \xmark & \cmark & \xmark & 55.4 & 85.2 & 85.6 & 76.1 & 85.8 & 86.2 & 76.7 & 56.1 & 85.4 & 76.8 & 56.1 & 85.5 & 75.9 \\
        PDA-R~\cite{pdar}& AAAI24 & \xmark & \cmark & \xmark & 55.4 & 85.1 & 85.8 & 75.2 & 85.2 & 85.2 & 74.2 & 55.2 & 85.8 & 74.7 & 55.8 & 86.3 & 75.3 \\
        DAMP-R~\cite{dampr}& CVPR24 & \xmark & \cmark & \xmark & 59.7 & 88.5 & 86.8 & 76.6 & 88.9 & 87.0 & 76.3 & 59.6 & 87.1 & 77.0 & 61.0 & 89.9 & 78.2 \\
        \midrule
        DIFO-V~\cite{difo} & CVPR24 & \cmark & \cmark & \xmark & 70.6 & 90.6 & 88.8 & 82.5 & 90.6 & 88.8 & 80.9 & 70.1 & 88.9 & \underline{83.4} & 70.5 & 91.2 & 83.1 \\
        ProDe-V~\cite{prode} & ICLR25 & \cmark & \cmark & \xmark & \underline{72.7} & \underline{92.3} & \underline{90.5} & \underline{82.5} & \underline{91.5} & \underline{90.7} & \underline{82.5} & \underline{72.5} & \underline{90.8} & 83.0 & \underline{72.6} & \underline{92.2} & \underline{84.5} \\

        \midrule
        \rowcolor{bestrow}
        \textbf{CoMA} & – & \cmark & \cmark & \cmark & \best{{83.8}} & \best{{95.6}} & \best{{94.2}} & \best{{89.1}} & \best{{95.9}} & \best{{93.7}} & \best{{89.2}} & \best{{83.9}} & \best{{93.6}} & \best{{89.4}} & \best{{84.6}} & \best{{95.8}} & \best{{90.7}} \\
        \bottomrule
        \end{tabular}
    }
\end{table*}

\textbf{Datasets.}
We conduct experiments on four widely used benchmarks for domain adaptation: 
{Office-31}~\cite{office31}, 
{Office-Home}~\cite{officehome}, 
{DomainNet-126}~\cite{domainnet-126}, 
and {VisDA-C}~\cite{visda-c}, which cover small-scale office environments, medium-scale diverse 
categories, large-scale multi-domain, and large-scale synthetic-to-real settings, 
respectively. Further statistics and details are provided in the \texttt{Supplementary}~\ref{supp:datasets}.

\vspace{4pt}
\noindent\textbf{Evaluation scenarios.}
We evaluate our method under three widely studied SFDA scenarios, following~\cite{shot}: 
the {Closed-set}, {Partial-set}, and {Open-set} settings. We conduct a comparative study between our \textbf{CoMA} and 18 representative baselines, categorized into three groups.
(1) Conventional SFDA methods without MFMs: SHOT~\cite{shot}, NRC~\cite{nrc}, GKD~\cite{gkd}, HCL~\cite{hcl}, AaD~\cite{aad}, AdaCon~\cite{adacon}, CoWA~\cite{cowa}, ELR~\cite{elr}, PLUE~\cite{plue}, CPD~\cite{cpd}, TPDS~\cite{tpds}.
(2) Single MFM-based UDA methods: DAPL-R~\cite{daplr}, PADCLIP-R~\cite{padclipr}, ADCLIP-R~\cite{adclipr}, PDA-R~\cite{pdar}, DAMP-R~\cite{dampr}, which serve as an informative upper bound with source data access.
(3) Single MFM-based SFDA methods: DIFO-V~\cite{difo} and ProDe-V~\cite{prode}, which employ a single MFM under the source-free setting.
For MFM-based methods, the suffixes -R and -V denote CLIP backbones with ResNet~\cite{resnet} and ViT~\cite{vit}, respectively. Following DIFO-V~\cite{difo} and ProDe-V~\cite{prode}, our CoMA uses ViT-B/32 as the CLIP backbone, and we also employ InstBLIP-XL~\cite{instblip}.

\begin{table}[t]
    \caption{{\bf Closed-set SFDA accuracy} (\%) on {Office-31}. The best accuracy is indicated in \textbf{bold} and the second best one is \underline{underlined}.}
    \vspace{-0.15cm}
    \label{tab:office31}
    \renewcommand\tabcolsep{1.8pt}
    \renewcommand\arraystretch{0.9}
    \scriptsize
    \centering
    \resizebox{\linewidth}{!}{
        \begin{tabular}{ l l | c c c c c c | c}
        \toprule
        Method & Venue & A$\rightarrow$D & A$\rightarrow$W & D$\rightarrow$A & D$\rightarrow$W & W$\rightarrow$A & W$\rightarrow$D & Avg. \\
        \midrule
        Source & – & 78.5 & 73.3 & 60.7 & 94.8 & 63.5 & 98.4 & 78.2 \\
        \midrule
        SHOT~\cite{shot} & ICML20 & 94.0 & 90.1 & 74.7 & 98.4 & 74.3 & 99.9 & 88.6 \\
        NRC~\cite{nrc} & NIPS21 & 96.0 & 90.8 & 75.3 & \underline{99.0} & 75.0 & \best{{100.}} & 89.4 \\
        GKD~\cite{gkd} & IROS21 & 94.6 & 91.6 & 75.1 & 98.7 & 75.1 & \best{{100.}} & 89.2 \\
        HCL~\cite{hcl} & NIPS21 & 94.7 & 92.5 & 75.9 & 98.2 & 77.7 & \best{{100.}} & 89.8 \\
        AaD~\cite{aad} & NIPS22 & 96.4 & 92.1 & 75.0 & \best{{99.1}} & 76.5 & \best{{100.}} & 89.9 \\
        AdaCon~\cite{adacon} & CVPR22 & 87.7 & 83.1 & 73.7 & 91.3 & 77.6 & 72.8 & 81.0 \\
        CoWA~\cite{cowa} & ICML22 & 94.4 & 95.2 & 76.2 & 98.5 & 77.6 & 99.8 & 90.3 \\
        ELR~\cite{elr} & ICLR23 & 93.8 & 93.3 & 76.2 & 98.0 & 76.9 & \best{{100.}} & 89.6 \\
        PLUE~\cite{plue} & CVPR23 & 89.2 & 88.4 & 72.8 & 97.1 & 69.6 & 97.9 & 85.8 \\
        CPD~\cite{cpd} & PR24 & 96.6 & 94.2 & 77.3 & 98.2 & 78.3 & \best{{100.}} & 90.8 \\
        TPDS~\cite{tpds} & IJCV24 & \underline{97.1} & 94.5 & 75.7 & 98.7 & 75.5 & 99.8 & 90.2 \\
        \midrule
        DIFO-V~\cite{difo} & CVPR24 & \best{{97.2}} & 95.5 & 83.0 & 97.2 & \underline{83.2} & 98.8 & 92.5 \\
        ProDe-V~\cite{prode} & ICLR25 & 96.8 & \best{{96.4}} & \underline{83.1} & 97.0 & 82.5 & 99.8 & \underline{92.6} \\
        \midrule
        \rowcolor{bestrow}
        \textbf{CoMA} & – & 95.6 & \underline{96.1} & \best{{86.8}} & 95.9 & \best{{86.5}} & 96.0 & \best{{92.8}} \\
        \bottomrule
        \end{tabular}
    }
\end{table}


\begin{table*}[!t]
    \caption{{\bf Closed-set SFDA accuracy} (\%) on {DomainNet-126} and {VisDA}. \textbf{SF}, \textbf{C} and \textbf{B} indicate source-free, CLIP and BLIP, respectively. The best accuracy is indicated in \textbf{bold} and the second best one is \underline{underlined}. 
    }
    \vspace{-0.15cm}
    \label{tab:DomainNet}
    \renewcommand\tabcolsep{3.8pt}
    \renewcommand\arraystretch{0.9}
    \scriptsize
    \centering
        \resizebox{\textwidth}{!}{
        \begin{tabular}{ l l | c c c | c c c c c c c c c c c c | c | c}
        \toprule
        \multirow{2}{*}{Method} &\multirow{2}{*}{Venue} 
        &\multirow{2}{*}{\textbf{SF}}
        &\multirow{2}{*}{\textbf{C}}
        &\multirow{2}{*}{\textbf{B}}
        &\multicolumn{13}{c}{\textbf{DomainNet-126}}\vline
        &{\textbf{VisDA}} \\
        & & & & & C$\to$P & C$\to$R & C$\to$S & P$\to$C & P$\to$R & P$\to$S & R$\to$C & R$\to$P & R$\to$S & S$\to$C & S$\to$P & S$\to$R & Avg. & Sy$\to$Re\\
        \midrule
        Source &-- &-- &-- &-- & 46.4 & 60.9 & 48.9 & 54.5 & 75.2 & 47.9 & 56.6 & 63.0 & 48.0 & 57.6 & 51.1 & 59.3 & 55.8 & 47.1\\
        \midrule
        SHOT~\cite{shot} & ICML20 & \cmark & \xmark & \xmark & 63.5 & 78.2 & 59.5 & 67.9 & 81.3 & 61.7 & 67.7 & 67.6 & 57.8 & 70.2 & 64.0 & 78.0 & 68.1 &82.9 \\
        NRC~\cite{nrc} & NIPS21 & \cmark & \xmark & \xmark & 62.6 & 77.1 & 58.3 & 62.9 & 81.3 & 60.7 & 64.7 & 69.4 & 58.7 & 69.4 & 65.8 & 78.7 & 67.5 & 85.9 \\
        GKD~\cite{gkd} & IROS21 & \cmark & \xmark & \xmark & 61.4 & 77.4 & 60.3 & 69.6 & 81.4 & 63.2 & 68.3 & 68.4 & 59.5 & 71.5 & 65.2 & 77.6 & 68.7 & 83.0 \\
        AdaCon~\cite{adacon} & CVPR22 & \cmark & \xmark & \xmark & 60.8 & 74.8 & 55.9 & 62.2 & 78.3 & 58.2 & 63.1 & 68.1 & 55.6 & 67.1 & 66.0 & 75.4 & 65.4 & 86.8\\
        CoWA~\cite{cowa} & ICML22 & \cmark & \xmark & \xmark & 64.6 & 80.6 & 60.6 & 66.2 & 79.8 & 60.8 & 69.0 & 67.2 & 60.0 & 69.0 & 65.8 & 79.9 & 68.6 & 86.9\\
        PLUE~\cite{plue} & CVPR23 & \cmark & \xmark & \xmark & 59.8 & 74.0 & 56.0 & 61.6 & 78.5 & 57.9 & 61.6 & 65.9 & 53.8 & 67.5 & 64.3 & 76.0 & 64.7 & 88.3\\
        TPDS~\cite{tpds} & IJCV24 & \cmark & \xmark & \xmark & 62.9 & 77.1 & 59.8 & 65.6 & 79.0 & 61.5 & 66.4 & 67.0 & 58.2 & 68.6 & 64.3 & 75.3 & 67.1 & 87.6\\
        \midrule
        DAPL-R~\cite{daplr} & TNNLS23 & \xmark & \cmark & \xmark & 72.4 & 87.6 & 65.9 & 72.7 & 87.6 & 65.6 & 73.2 & 72.4 & 66.2 & 73.8 & 72.9 & 87.8 & 74.8 & 86.9\\
        ADCLIP-R~\cite{adclipr} & ICCVW23 & \xmark & \cmark & \xmark & 71.7 & 88.1 & 66.0 & 73.2 & 86.9 & 65.2 & 73.6 & 73.0 & 68.4 & 72.3 & 74.2 & 89.3 & 75.2 & 87.7\\
        DAMP-R~\cite{dampr} & CVPR24 & \xmark & \cmark & \xmark & 76.7 & 88.5 & 71.7 & 74.2 & 88.7 & 70.8 & 74.4 & 75.7 & 70.5 & 74.9 & 76.1 & 88.2 & 77.5 & 88.4\\
        \midrule
        DIFO-V~\cite{difo} & CVPR24 & \cmark & \cmark & \xmark & 76.6 & 87.2 & 74.9 & 80.0 & 87.4 & 75.6 & 80.8 & 77.3 & 75.5 & 80.5 & 76.7 & 87.3 & 80.0 & 90.3\\
        ProDe-V~\cite{prode} & ICLR25 & \cmark & \cmark & \xmark & \underline{83.2} & \best{{92.4}} & \underline{79.0} & \underline{85.0} & \best{{92.3}} & \underline{79.3} & \underline{85.5} & \underline{83.1} & \underline{79.1} & \underline{85.5} & \underline{83.4} & \best{{92.4}} & \underline{85.0} & \underline{91.0}\\
        \midrule
        \rowcolor{bestrow}
        \textbf{CoMA} & -- & \cmark & \cmark & \cmark & \best{{84.8}} & \underline{90.7} & \best{{84.8}} & \best{{89.2}} & \underline{90.8} & \best{{85.3}} & \best{{89.2}} & \best{{85.9}} & \best{{85.4}} & \best{{88.9}} & \best{{85.3}} & \underline{90.4} & \best{{87.6}} & \best{92.5}\\
        \bottomrule
        \end{tabular}
    }
\end{table*}

\subsection{Comparison Result}
\noindent\textbf{Comparison results on Closed-set SFDA.}
Tables~\ref{tab:officehome}--\ref{tab:DomainNet} report quantitative comparisons on four benchmarks under the closed-set SFDA setting. 
On Office-31, CoMA achieves 92.8\% average accuracy, surpassing the previous best method ProDe-V by +\textbf{0.2}\%.
On Office-Home, CoMA attains 90.7\%, improving upon ProDe-V by +\textbf{6.2}\%, demonstrating the advantage of dual-MFM guidance under complex domain diversity.
On DomainNet-126, CoMA reaches 87.6\%, outperforming ProDe-V by +\textbf{2.6}\% and verifying strong scalability to large label spaces.
Finally, on VisDA, CoMA achieves 92.5\%, exceeding ProDe-V by +\textbf{1.5}\%, with large improvements in categories involving complex multi-object scenes such as \emph{car} (+4.6\%) and \emph{person} (+7.2\%).
Overall, our CoMA establishes new state-of-the-art performance on all four benchmarks.

\vspace{4pt}
\noindent\textbf{Comparison results on Partial-set and Open-set SFDA.}
Table~\ref{tab:po-set} reports comparison results on {Partial-set} and {Open-set} SFDA. 
Our method achieves clear improvements across both settings, surpassing previous best results by a large margin. Specifically, it outperforms ProDe-V~\cite{prode} by +\textbf{7.3}\% on Partial-set and +\textbf{2.8}\%  on Open-set, demonstrating strong robustness to missing or unseen target classes.

\begin{table}[t]
    \caption{{\bf Partial-set and Open-set SFDA} (\%) on {Office-Home}. The best accuracy is indicated in \textbf{bold} and the second best one is \underline{underlined}. 
    }
    \vspace{-0.15cm}
    \label{tab:po-set}
    \renewcommand\tabcolsep{4.0pt}
    \renewcommand\arraystretch{0.9}
    \scriptsize
    \centering
    \resizebox{\linewidth}{!}{
        \begin{tabular}{llc|llc}
        \toprule
        Partial-set  &Venue &Avg. &Open-set &Venue & Avg.\\
        \midrule
        Source  &--  &62.8&Source &--& 46.6 \\
        \midrule
        SHOT~\cite{shot} &ICML20 &79.3 & SHOT~\cite{shot} &ICML20 &72.8 \\
        HCL~\cite{hcl} &NIPS21 &79.6 &HCL~\cite{hcl} &NIPS21 &72.6 \\
        AaD~\cite{aad} &NIPS22 &79.7 &AaD~\cite{aad} &NIPS22 &71.8 \\
        CoWA~\cite{cowa} &ICML22 &83.2 &CoWA~\cite{cowa} &ICML22 &73.2 \\
        CRS~\cite{crs} &CVPR23 &80.6 &CRS~\cite{crs} &CVPR23 &73.2 \\
        \midrule
        DIFO-V~\cite{difo} &CVPR24 &84.1 &DIFO-V~\cite{difo} &CVPR24 &75.9 \\
        ProDe-V~\cite{prode} &ICLR25 &\underline{84.2} &ProDe-V~\cite{prode} &ICLR25 &\underline{82.6} \\
        \midrule
        \rowcolor{bestrow}
        \textbf{CoMA} & -- & \best{{91.5}} &\textbf{CoMA} &-- &\best{{85.4}}\\
        \bottomrule
        \end{tabular} 
    }
\end{table}

\subsection{Ablation Studies}
\noindent\textbf{Effect of the components.}
We conduct an ablation to evaluate the contribution of each loss component in our framework, as shown in Table~\ref{tab:ablation}. 
Starting from the plain source model baseline (61.6\%), introducing $\mathcal{L}_{\mathrm{SIM}}$ substantially improves performance to 83.0\%, confirming its effectiveness in enhancing target discriminability with MFMs' guidance. 
When $\mathcal{L}_{\mathrm{AGS}}$ is applied, the accuracy rises to 90.4\%, showing the benefit of supervision derived from MFMs agreement. 
Adding $\mathcal{L}_{\mathrm{MC}}$ further increases the average to 91.5\%, indicating the importance of task adaptation.
Finally, incorporating $\mathcal{L}_{\mathrm{CD}}$ yields the best result of 92.0\%, with +30.4\% gain over the baseline. 
The results demonstrate a clear cumulative effect, leading to robust adaptation.

\vspace{4pt}
\noindent\textbf{Information objective comparison.}
To verify the effectiveness of our proposed DMI, 
we compare it with KL divergence and standard MI on two benchmarks with different class scales: Office-Home ($K=65$) and VisDA ($K=12$). 
As shown in Table~\ref{tab:ablation2}, replacing KL with MI brings notable improvement by modeling richer dependencies between models in unsupervised settings. 
Further introducing DMI yields consistent gains for both ProDe~\cite{prode} and our CoMA, with +1.1\% average increases over MI for both.
This confirms that DMI provides more stable and discriminative alignment, effectively mitigating inaccurate estimation in high-class scenarios such as Office-Home, while also improving under low-class settings like VisDA.

\begin{table}[t]
    \caption{{\bf Effect of the components} (\%) on {Office-31}, {Office-Home} and {VisDA} under the closed-set SFDA setting.}
    \vspace{-0.15cm}
    \label{tab:ablation}
    \renewcommand\tabcolsep{4.0pt}
    \renewcommand\arraystretch{0.9}
    \scriptsize
    \centering
    \resizebox{\linewidth}{!}{
        \begin{tabular}{cccc|ccc|c}
        \toprule
        {$\mathcal{L}_{\mathrm{CD}}$} &{$\mathcal{L}_{\mathrm{MC}}$} &{$\mathcal{L}_{\mathrm{AGS}}$} & {$\mathcal{L}_{\mathrm{SIM}}$}&{\bf Office-31} &{\bf Office-Home} &{\bf VisDA} &{Avg.} \\
        \midrule
        \xmark  &\xmark  &\xmark &\xmark &{78.2} &{59.6} &{47.1} &{61.6} \\
        \xmark  &\xmark  &\xmark &\cmark &{90.0} &{84.8} &{74.1} &{83.0} \\
        \xmark  &\xmark  &\cmark &\cmark &{90.9} &{89.2} &{91.1} &{90.4} \\
        \xmark  &\cmark  &\cmark &\cmark &{91.8} &{90.5} &{92.3} &{91.5} \\
        \midrule
        \rowcolor{bestrow}
        \cmark  &\cmark  &\cmark &\cmark &\best{{92.8}} &\best{{90.7}}&\best{{92.5}} &\best{{92.0}} \\
        \bottomrule
        \end{tabular} 
    }
\end{table}

\begin{table}[t]
    \caption{
    {\bf Effect of our DMI} (\%) on {Office-Home} and {VisDA} under the closed-set SFDA setting.
    }
    \vspace{-0.15cm}
    \label{tab:ablation2}
    \renewcommand\tabcolsep{4.0pt}
    \renewcommand\arraystretch{0.9}
    \scriptsize
    \centering
    \resizebox{\linewidth}{!}{
        \begin{tabular}{ll|cc|c}
        \toprule
        Method  &Venue &{\bf Office-Home} &{\bf VisDA} & Avg.\\
        \midrule
        ProDe-V~\cite{prode} (w/ KL) & ICLR25 &72.9& 89.8 &81.4\\
        ProDe-V~\cite{prode} (w/ MI) & ICLR25 &84.5& 91.0 &87.8\\
        \rowcolor{bestrow}
        ProDe-V~\cite{prode} (w/ DMI)& ICLR25 &\best{85.6} &\best{91.8} &\best{88.7}\\
        \midrule
        \textbf{CoMA} (w/ KL) &--&87.6 &89.0 &88.3\\
        \textbf{CoMA} (w/ MI)&--&88.8 &92.1 &90.5\\
        \rowcolor{bestrow}
        \textbf{CoMA} (w/ DMI) & -- &\best{{90.7}}&\best{{92.5}} &\best{{91.6}}\\
        \bottomrule
        \end{tabular}
    }
\end{table}

\begin{table}[t]
    \caption{
    {\bf Reliance analysis} (\%) on {Office-Home} and {VisDA} under the closed-set SFDA setting.
    The table consists of four blocks:  
    (1) existing MFM-based SFDA methods,  
    (2) zero-shot MFMs,  
    (3) our CoMA with CLIP-V32-based configurations, and  
    (4) our CoMA with CLIP-V16-based configurations. Here, V32 and V16 denote ViT-B/32 and ViT-B/16.
    }
    \vspace{-0.15cm}
    \label{tab:reliance}
    \renewcommand\tabcolsep{4.0pt}
    \renewcommand\arraystretch{0.9}
    \scriptsize
    \centering
    \resizebox{\linewidth}{!}{
        \begin{tabular}{ll|cc|c}
        \toprule
        Method  &Venue &{\bf Office-Home} &{\bf VisDA} & Avg.\\
        \midrule
        DIFO-V32~\cite{difo} &CVPR24 &83.1& 90.3 &86.7\\
        DIFO-V16~\cite{difo} &CVPR24 &85.5& 91.0 &88.3\\
        ProDe-V32~\cite{prode} &ICLR25 &84.5& 91.0 &87.8\\
        ProDe-V16~\cite{prode} &ICLR25 &86.9& 91.7 &89.3\\
        \midrule
        CLIP-V32~\cite{clip} &ICML21 &79.1& 87.1 &83.1\\
        CLIP-V16~\cite{clip} &ICML21 &81.8& 88.9 &85.4\\
        LLaVA-7B~\cite{llava} &NIPS23 &78.8 &88.5 & 83.7\\
        BLIP2-XL~\cite{blip2} &ICML23 &84.6& 82.2 &83.4\\
        InstBLIP-XL~\cite{instblip} &NIPS24 &85.2& 86.8 &86.0\\
        \midrule
        \textbf{CoMA} (only CLIP-V32) & -- &84.5&91.7 &88.1\\
        \textbf{CoMA} (CLIP-V32, LLaVA-7B) & -- &87.7 &92.7 &90.2\\ 
        \textbf{CoMA} (CLIP-V32, BLIP2-XL) & -- &90.8 &91.8 &91.3\\ 
        \rowcolor{bestrow}
        \textbf{CoMA} (CLIP-V32, InstBLIP-XL) & -- &90.7 & 92.5 &\best{91.6}\\
        \midrule
        \textbf{CoMA} (only CLIP-V16) & -- &86.8 &92.3 &89.6\\
        \textbf{CoMA} (CLIP-V16, LLaVA-7B) & -- &88.3 &{92.9} &90.7\\ 
        \textbf{CoMA} (CLIP-V16, BLIP2-XL) & -- &{91.4} &92.1 &{91.7}\\
        \rowcolor{bestrow}
        \textbf{CoMA} (CLIP-V16, InstBLIP-XL) & -- &{{91.4}} &{{92.8}} &\best{{92.1}}\\
        \bottomrule
        \end{tabular}
    }
\end{table}

\begin{figure*}[t]
    \centering
    \begin{subfigure}[t]{0.15\linewidth}
        \centering
        \includegraphics[width=\linewidth]{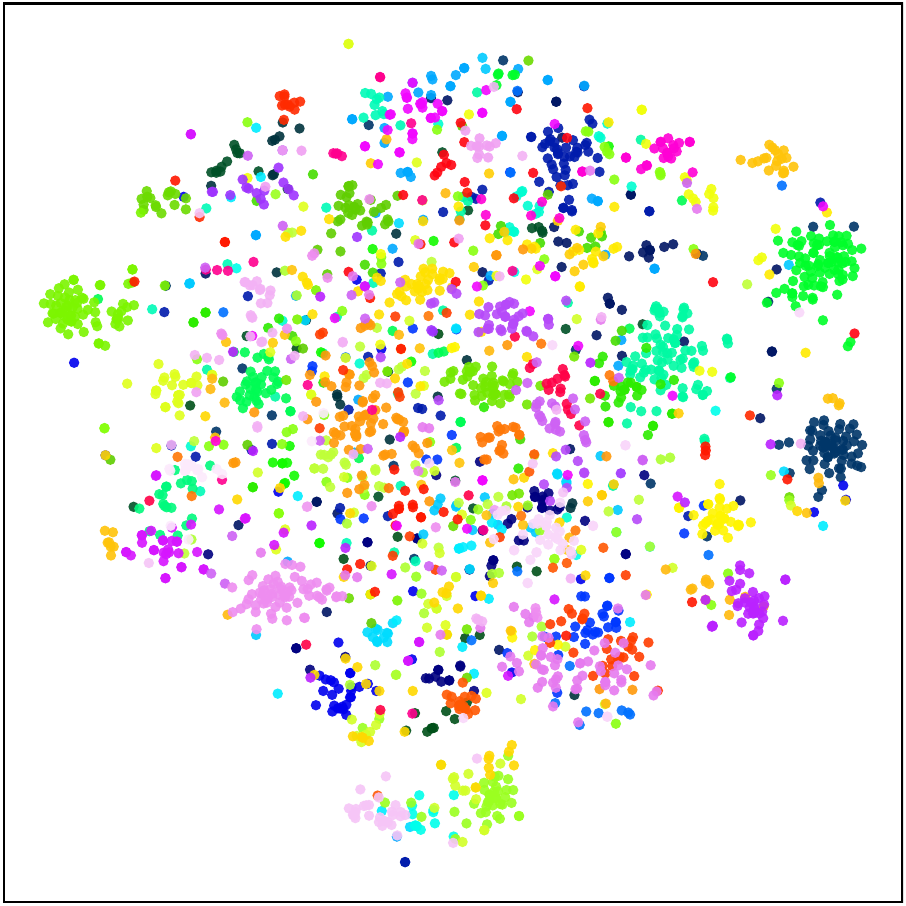}
        \caption*{\small Source}
    \end{subfigure}
    \hspace{-1pt}
    \begin{subfigure}[t]{0.15\linewidth}
        \centering
        \includegraphics[width=\linewidth]{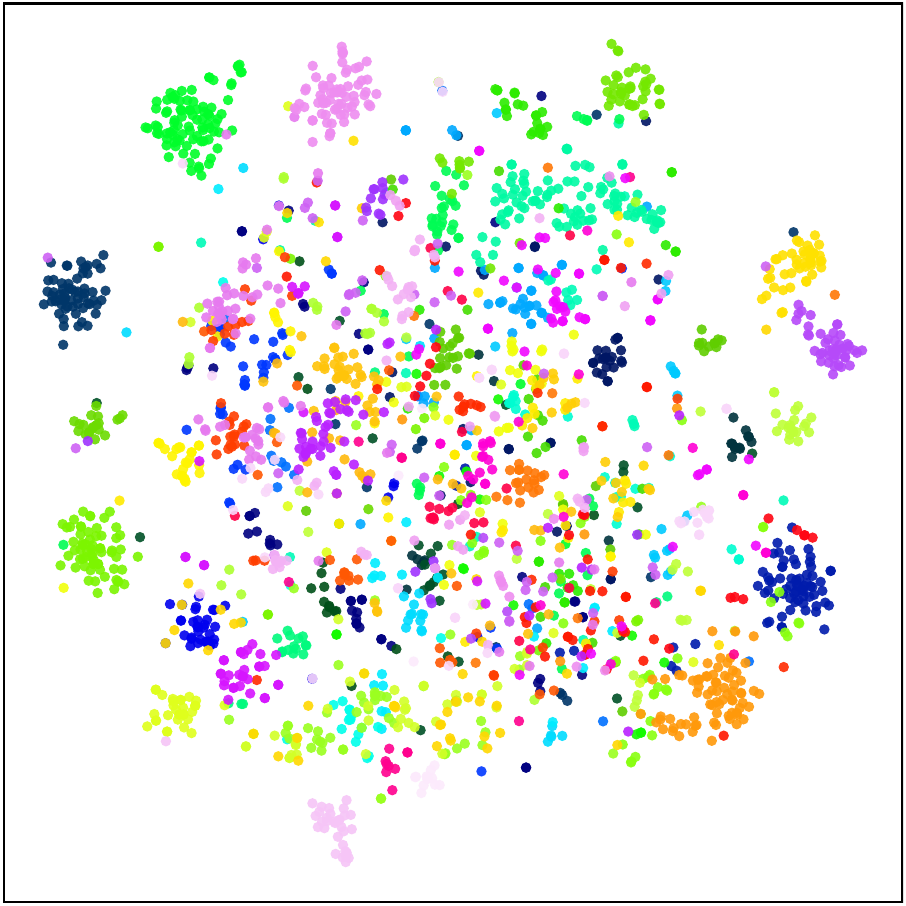}
        \caption*{\small CLIP-V}
    \end{subfigure}
    \hspace{-1pt}
    \begin{subfigure}[t]{0.15\linewidth}
        \centering
        \includegraphics[width=\linewidth]{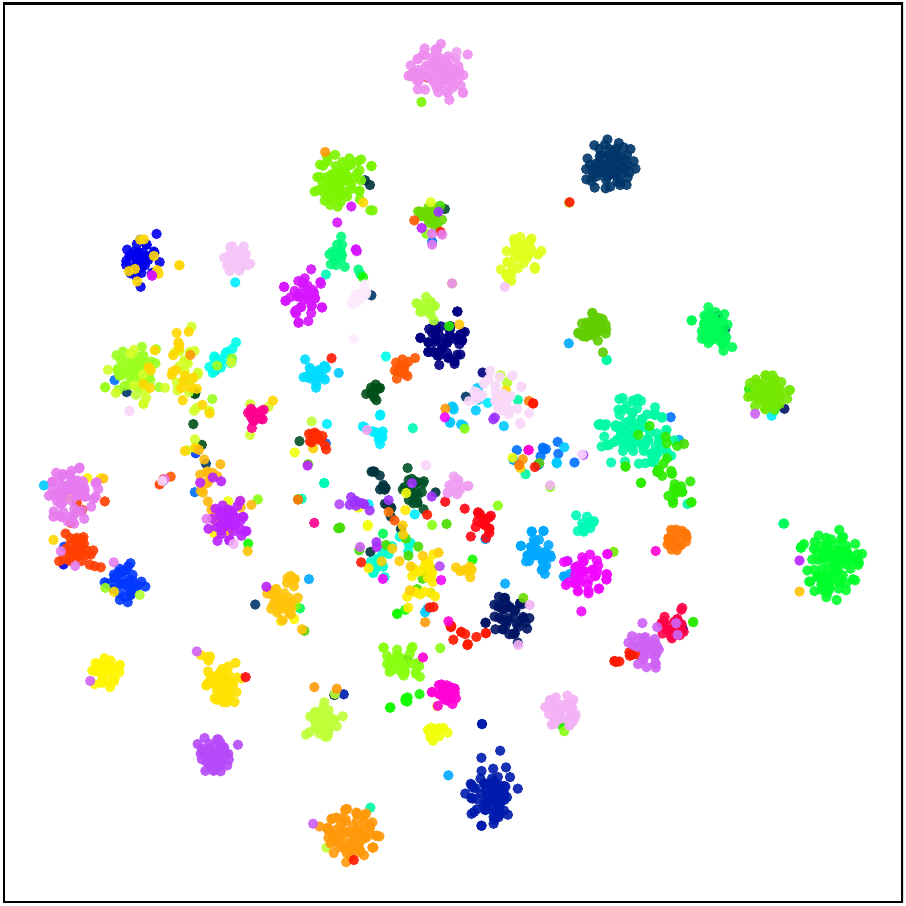}
        \caption*{\small BLIP-Proxy}
    \end{subfigure}
    \hspace{-1pt}
    \begin{subfigure}[t]{0.15\linewidth}
        \centering
        \includegraphics[width=\linewidth]{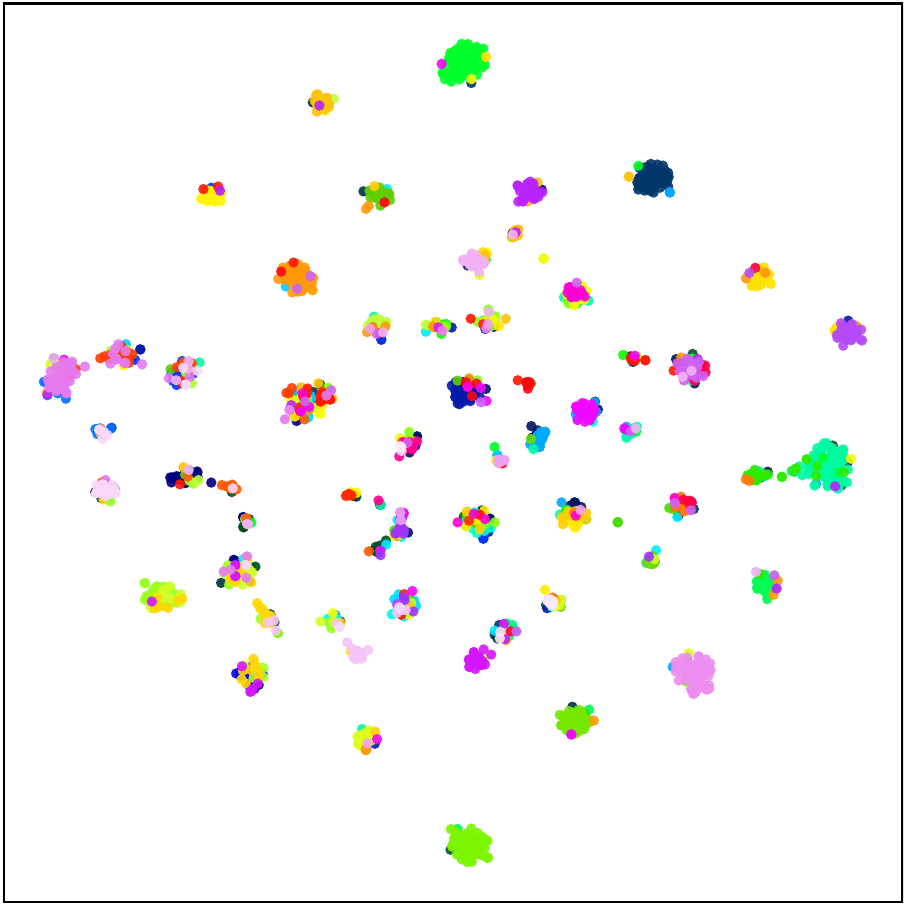}
        \caption*{\small SHOT}
    \end{subfigure}
    \hspace{-1pt}
    \begin{subfigure}[t]{0.15\linewidth}
        \centering
        \includegraphics[width=\linewidth]{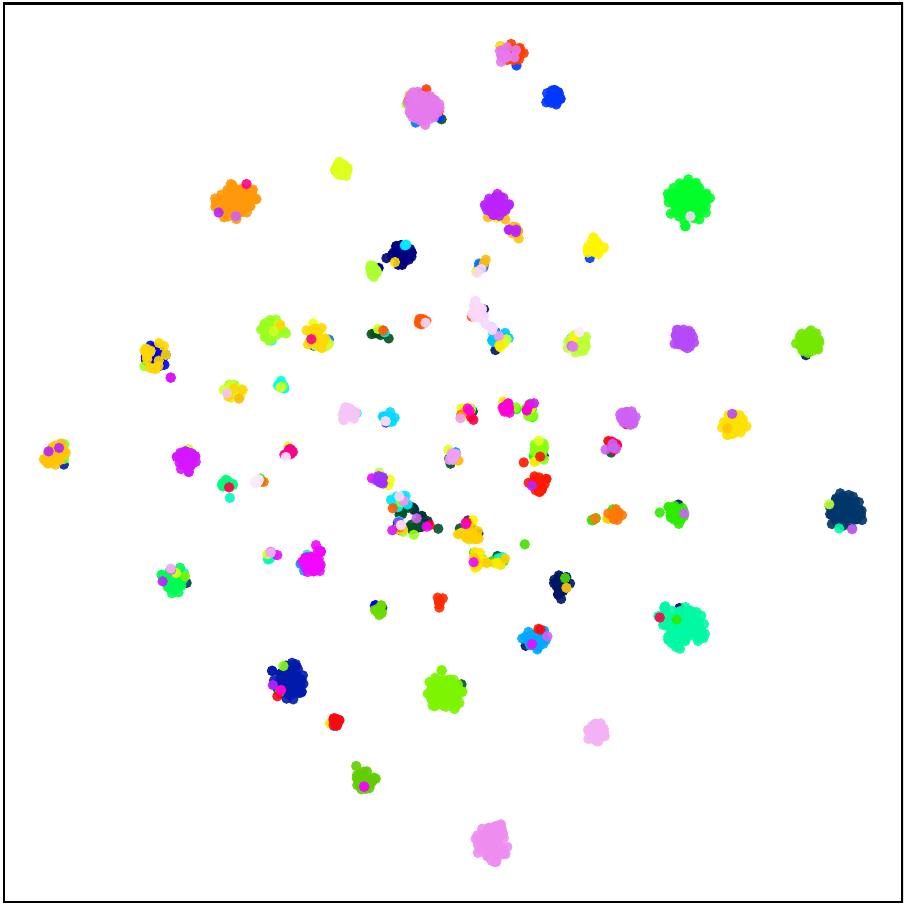}
        \caption*{\small ProDe-V}
    \end{subfigure}
    \hspace{-1pt}
    \begin{subfigure}[t]{0.15\linewidth}
        \centering
        \includegraphics[width=\linewidth]{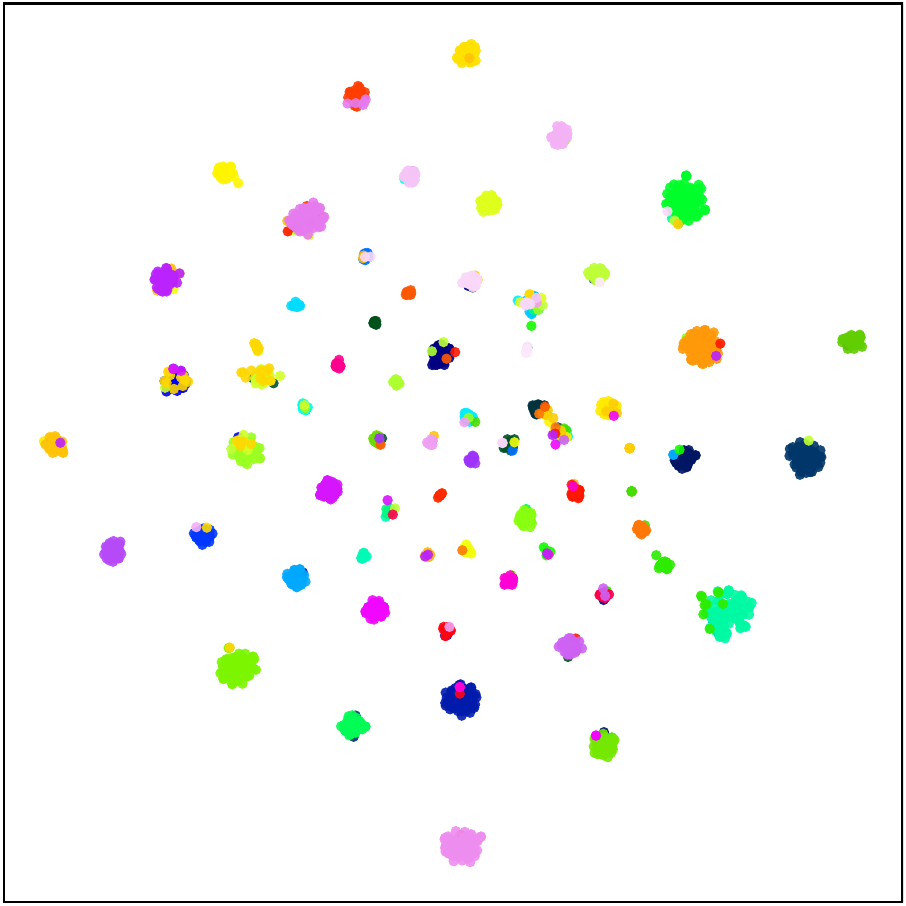}
        \caption*{\small CoMA}
    \end{subfigure}

    \caption{
    t-SNE feature visualization on the transfer task Cl$\to$Ar in {Office-Home}. Each color corresponds to one of the 65 object categories.
    }
    \label{fig:tsne}
    \vspace{-0.2cm}
\end{figure*}

\subsection{Additional Analysis}
\noindent\textbf{Feature visualization.}
Fig.~\ref{fig:tsne} shows the t-SNE~\cite{tsne} visualization results on the Office-Home Cl$\to$Ar transfer task.
To analyze the discriminability of target-domain embeddings, we compare the feature distributions of the source model, CLIP-V, BLIP-Proxy, SHOT~\cite{shot}, ProDe-V~\cite{prode}, and our CoMA.
The source model produces highly entangled feature distributions, showing that target samples from different classes are poorly separated.
CLIP-V provides a more structured layout but still suffers from fuzzy class transitions, since it prioritizes category-level alignment over fine-grained visual cues. 
In contrast, BLIP-Proxy trained on BLIP-generated pseudo-labels shows improved local compactness, reflecting the fine-grained supervision induced by BLIP's guidance. Compared to SHOT~\cite{shot} and ProDe-V~\cite{prode}, our CoMA yields the most balanced distribution with notably more compact intra-class clusters and clearer inter-class separation, achieving both global alignment and local discriminability.

\vspace{4pt}
\noindent\textbf{Reliance analysis.}
Table~\ref{tab:reliance} evaluates the effectiveness of CoMA across different MFM configurations and CLIP backbones. We pair CLIP~\cite{clip} with diverse MFMs including BLIP2-XL~\cite{blip2}, InstBLIP-XL~\cite{instblip}, and LLaVA-7B~\cite{llava} to verify generality beyond specific architectures. 
Compared with existing SFDA methods DIFO~\cite{difo} and ProDe~\cite{prode}, standalone MFMs remain domain-agnostic with limited zero-shot performance. When integrated into our framework, CoMA yields substantial improvements across all configurations. 
Among various pairings, CLIP with InstBLIP-XL achieves the best result across both CLIP backbones.
Remarkably, CoMA using only CLIP with separately trained CLIP-proxy outperforms ProDe under both backbones. While this effect is weaker than leveraging two complementary MFMs, this still indicates that CoMA benefits from distinct learning trajectories that introduce mild yet meaningful semantic diversity. Additionally, upgrading CLIP's backbone from V32 to V16 consistently improves performance across all MFM pairs, confirming that CoMA is robust and not sensitive to CLIP's backbone.

\begin{figure}[t]
    \centering
    \begin{subfigure}[t]{0.47\linewidth}
        \centering
        \includegraphics[width=\linewidth]{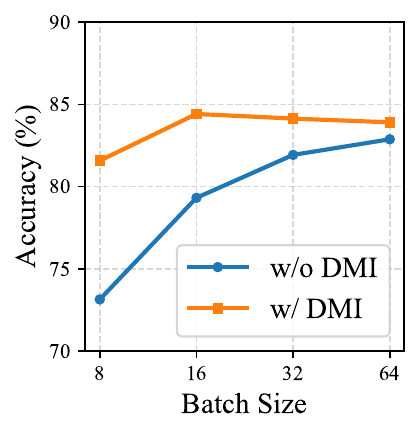}
    \end{subfigure}
    \hspace{-1pt}
    \begin{subfigure}[t]{0.47\linewidth}
        \centering
        \includegraphics[width=\linewidth]{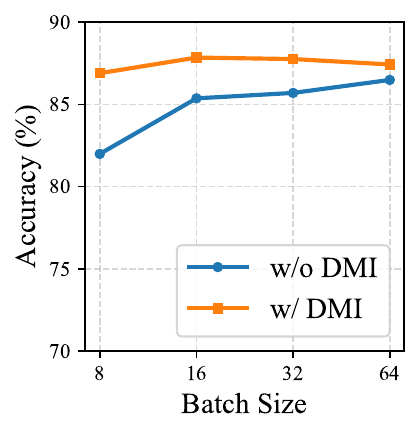}
    \end{subfigure}
    \vspace{-0.2cm}
    \caption{
    Batch size sensitivity of DMI on the transfer task Cl$\to$Ar in Office-Home. 
    \textbf{Left}: ProDe-V~\cite{prode}. \textbf{Right}: Our CoMA. DMI maintains stable performance across batch sizes, effectively mitigating the degradation observed with the standard MI.
    }
    \label{fig:batch}
    \vspace{-0.2cm}
\end{figure}

\vspace{4pt}
\noindent\textbf{Batch size sensitivity of DMI.}
To verify DMI's robustness under mini-batch training, we evaluate performance across varying batch sizes on Office-Home Cl$\to$Ar. 
As shown in Fig.~\ref{fig:batch}, both ProDe-V~\cite{prode} and our CoMA with standard MI exhibit severe degradation as batch size decreases, demonstrating the instability caused by false dependencies under incomplete class coverage within mini-batches. 
In comparison, applying DMI effectively mitigates this degradation issue, achieving superior performance over standard MI across most batch sizes. This confirms that DMI successfully addresses the instability of conventional MI, providing stable adaptation under mini-batch training.

%% file: sec/5_conclusion.tex
\section{Conclusion}

We propose CoMA, a novel SFDA framework that jointly leverages two distinctive MFMs with complementary semantics. Through bidirectional adaptation, MFMs and the target model collaborate synergistically: TCA harmonizes MFMs with the target domain while preserving their distinct semantics, and MDA distills 
this complementary knowledge for robust target adaptation. Since mini-batch training produces false dependencies due to incomplete class coverage, we introduce DMI to address this by decomposing 
MI into confident and uncertain regions. Extensive experimental results across four benchmarks verify 
that multi-foundation collaboration provides a principled and effective paradigm for knowledge-driven SFDA.

%% file: sec/6_supplementary.tex
\clearpage
\setcounter{page}{1}
\maketitlesupplementary

\section{Proof of Proposition 1}
\label{supp:proof}
\noindent{\bf \textit{Proof.}}
We first recall two following well-known properties of MI~\cite{kl}:  
(1) non-negativity and (2) boundedness by entropy, where $K$ is the number of classes.
\begin{equation*}
    \begin{split}
        (1) \quad I(X;Y) &= D_{\mathrm{KL}}(p(x,y)\,\|\,p(x)p(y)) \ge 0,\\
        (2) \quad I(X;Y) &\le \min\{H(X),H(Y)\} \le \log K.
    \end{split}
\end{equation*}

Following Eq. (2) in the main paper, $\mathcal{S}$ denotes the candidate class subset,
and $\mathcal{S}^{\complement}$ denotes its complement within the class space $\mathcal{C}$,
such that $\mathcal{S} \cup \mathcal{S}^{\complement} = \mathcal{C}$ and $\mathcal{S} \cap \mathcal{S}^{\complement} = \emptyset$.
According to Property (1), the MI computed within each region is non-negative:
\begin{equation*}
I(X_\mathcal{S};Y_\mathcal{S}) \ge 0,
\quad
I(X_{\mathcal{S}^{\complement}};Y_{\mathcal{S}^{\complement}}) \ge 0,
\end{equation*}

\noindent
From Property (2), each term is upper-bounded by the entropy of its class support:
\begin{equation*}
    \begin{split}
    I(X_\mathcal{S};Y_\mathcal{S})
    &\le \min\{H(X_\mathcal{S}), H(Y_\mathcal{S})\} \le \log |\mathcal{S}|,\\
    I(X_{\mathcal{S}^{\complement}};Y_{\mathcal{S}^{\complement}})
    &\le \min\{H(X_{\mathcal{S}^{\complement}}), H(Y_{\mathcal{S}^{\complement}})\}
    \le \log |\mathcal{S}^{\complement}|,
    \end{split}
\end{equation*}

\noindent
Thus, we obtain the following two inequalities:
\begin{equation*}
    \begin{split}
        0 &\le I(X_\mathcal{S};Y_\mathcal{S}) \le \log|\mathcal{S}|\\
        0 &\le I(X_{\mathcal{S}^{\complement}};Y_{\mathcal{S}^{\complement}})\le\log|\mathcal{S}^{\complement}|.
    \end{split}
\end{equation*}

\noindent
Substituting these bounds into the definition of DMI:
\begin{equation*}
I_D(X;Y)
= I(X_\mathcal{S};Y_\mathcal{S})
	-\frac{\log|\mathcal{S}|}{\log|\mathcal{S}^{\complement}|} \cdot I(X_{\mathcal{S}^{\complement}};Y_{\mathcal{S}^{\complement}}),
\end{equation*}

\noindent
we finally have:
\begin{equation*}
-\log|\mathcal{S}| \le I_D(X;Y) \le \log|\mathcal{S}|.
\end{equation*}

Therefore, $I_D(X;Y)$ remains bounded by $\log|\mathcal{S}|$,
ensuring stable scaling with respect to the candidate class subset size. This completes the proof. $\square$

\vspace{4pt}
\noindent{\bf \textit{Remark.}}
The logarithmic ratio $\frac{\log|\mathcal{S}|}{\log|\mathcal{S}^{\complement}|}$ is positive whenever both $\mathcal{S}$ and $\mathcal{S}^{\complement}$ contain at least two distinct class elements.
In practice, for numerical stability, when $|\mathcal{S}^{\complement}| \le 1$,
we set the ratio to zero and exclude the corresponding uncertain region from optimization.
Similarly, when $|\mathcal{S}| \le 1$, indicating an insufficient or degenerate confident joint subset,
the entire batch is skipped to avoid unstable or numerically meaningless updates during training.

\section{Extended Definition of DMI}
\label{supp:extended_dmi}

In the main paper, Definition~\ref{def:dmi} introduces DMI without an additional scale parameter to emphasize the conceptual clarity of its core mechanism: enhancement of confident dependencies and suppression of uncertain ones. Since DMI is a novel formulation, we present it in its simplest form for theoretical analysis.

In this section, we extend the definition with an tunable scale parameter $\lambda$ for two purposes: (1) analyzing the relative contribution of enhancement and suppression terms, and (2) empirically validating the optimal balance for practical deployment. This enables systematic investigation of the enhancement-to-suppression ratio and facilitates optimization tuning across different datasets.

\vspace{4pt}
\noindent\textbf{Extended DMI formulation.}
We introduce the extended DMI with tunable scale parameter $\lambda > 0$:
\begin{equation*}
I_D^{(\lambda)}(X;Y) = I(X_\mathcal{S};Y_\mathcal{S}) 
- \lambda \cdot \frac{\log|\mathcal{S}|}{\log|\mathcal{S}^{\complement}|} 
\cdot I(X_{\mathcal{S}^{\complement}};Y_{\mathcal{S}^{\complement}}),
\label{eq:dmi_extended}
\end{equation*}
where $\lambda$ controls the relative strength of suppression with respect to enhancement. Definition~\ref{def:dmi} corresponds to the case $\lambda = 1$.

\vspace{4pt}
\noindent\textbf{Interpretation.}
The balancing factor $\frac{\log|\mathcal{S}|}{\log|\mathcal{S}^{\complement}|}$ normalizes the suppression term to the same scale as enhancement. Thus, $\lambda$ represents the enhancement-to-suppression ratio on a normalized scale, invariant to subset sizes. For example, $\lambda = 0.5$ means suppression operates at 50\% of enhancement strength, regardless of batch composition.

\vspace{4pt}
\noindent\textbf{Theoretical properties.}
The extended formulation inherits all properties of Definition~\ref{def:dmi}. The bounded condition in Proposition~\ref{propose:DMI} generalizes to:
\begin{equation*}
-\lambda \log|\mathcal{S}| \le I_D^{(\lambda)}(X;Y) \le \log|\mathcal{S}|,
\end{equation*}
recovering Proposition 1 when $\lambda = 1$. The asymmetric bound reflects DMI's design: full-strength enhancement (upper bound $\log|\mathcal{S}|$) and controlled suppression (lower bound $-\lambda\log|\mathcal{S}|$).

\begin{figure}[t]
    \centering
    \begin{subfigure}[t]{0.48\linewidth}
        \centering
        \includegraphics[width=\linewidth]{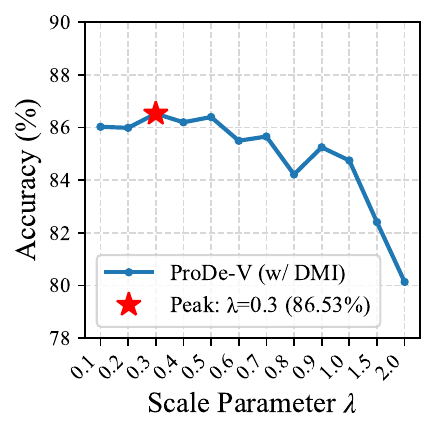}
    \end{subfigure}
    \hspace{-1pt}
    \begin{subfigure}[t]{0.48\linewidth}
        \centering
        \includegraphics[width=\linewidth]{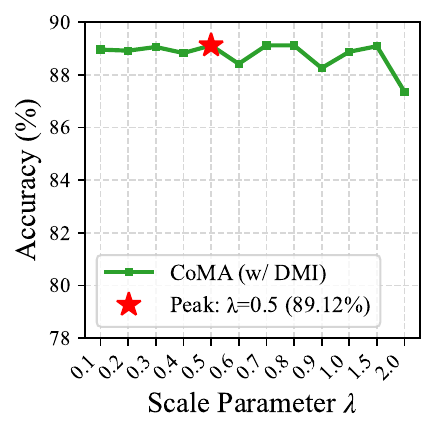}
    \end{subfigure}
    \vspace{-0.2cm}
    \caption{
    Scale $\lambda$ sensitivity of extended DMI on the transfer task Cl$\to$Ar in Office-Home. 
    \textbf{Left}: ProDe-V~\cite{prode}. \textbf{Right}: Our CoMA.
    }
    \label{fig:lambda}
    \vspace{-0.2cm}
\end{figure}

\vspace{4pt}
\noindent\textbf{Sensitivity analysis of $\lambda$.}
To analyze the relative contribution of enhancement and suppression terms and find the optimal balance, we conduct ablation on the DMI scale parameter $\lambda \in [0.1, 2.0]$ on Office-Home Cl$\to$Ar with batch size 32. Here, $\lambda<1$ emphasizes enhancement over suppression, and $\lambda>1$ applies stronger suppression. As shown in Fig.~\ref{fig:lambda}, CoMA achieves peak performance at $\lambda=0.5$, while ProDe-V~\cite{prode} peaks at $\lambda=0.3$.  Interestingly, the performance remains stable even at low $\lambda$ values, suggesting that the primary benefit of DMI stems from selective enhancement within $\mathcal{S}$, while suppression of $\mathcal{S}^{\complement}$ plays a secondary role. However, excessive suppression ($\lambda = 2.0$) causes significant degradation, showing that over-penalization disrupts the optimization balance.

\section{Algorithm Detail of Our CoMA}
Algorithm~\ref{alg:coma} presents the training procedure of CoMA, which alternates between TCA (lines 5--8) and MDA (lines 10--15) stages as described in Section 3 of the main paper.

\begin{algorithm}[H]
    \caption{Training procedure of CoMA}
    \label{alg:coma}
    \small
    \noindent\textbf{Input}: Unlabeled target data $\mathcal{X}_t$, pre-trained source model $\theta_s$, 
    CLIP $\Theta_c$, learnable prompt context $v$, BLIP-proxy $\theta_b$, max epochs $E$, hyper-parameters $\alpha, \beta$.\\
    \noindent\textbf{Output}: Adapted target model $\theta_t$.\\
    \textbf{Procedure}:
    \begin{algorithmic}[1]
    \STATE \textbf{Initialization}: $\theta_t \leftarrow \theta_s$ and $v\leftarrow$ ``a photo of a".
    \FOR{epoch $= 1$ to $E$}
        \FOR{each mini-batch $\mathcal{B} \sim \mathcal{X}_t$}
            \STATE \textcolor{gray}{============ Stage 1: TCA ============}
            \STATE {Forward} and {obtain} predictions: \\
            $p_c \leftarrow \Theta_c(\mathcal{B}, v)$, 
            $p_b \leftarrow \theta_b(\mathcal{B})$, $p_t \leftarrow \theta_t(\mathcal{B})$.
            
            \STATE Compute $\mathcal{L}_{\text{MC}}$ (Eq.~\eqref{eq:mc}).
            \STATE Compute $\mathcal{L}_{\text{CD}}$ (Eq.~\eqref{eq:cd}).
            
            \STATE Update $v$ and $\theta_b$ by optimizing 
            $\mathcal{L}_{\text{TCA}}$ (Eq.~\eqref{eq:tca}).
            
            \STATE \textcolor{gray}{============ Stage 2: MDA ============}
            \STATE Re-compute MFM predictions:\\
            $p_c \leftarrow \Theta_c(\mathcal{B}, v)$, $p_b \leftarrow \theta_b(\mathcal{B})$.
            
            \STATE Obtain pseudo-labels $\hat{y}_t$ from MFM agreement: \\
            $\hat{y}_t=k$~~if $\arg\max_k~p_c=\arg\max_k~p_b$,~~else ignore.
            \STATE Obtain candidate class subset from $p_c$ and $p_b$ (Eq.~\eqref{eq:ccs}).
            \STATE Compute $\mathcal{L}_{\text{AGS}}$ (Eq.~\eqref{eq:ags}).
            \STATE Compute $\mathcal{L}_{\text{SIM}}$ (Eq.~\eqref{eq:sim}).
            
            \STATE Update $\theta_t$ by optimizing $\mathcal{L}_{\text{MDA}}$ (Eq.~\eqref{eq:mda}).
        \ENDFOR
    \ENDFOR
    
    \RETURN Adapted target model $\theta_t$
    \end{algorithmic}
\end{algorithm}

\section{Evaluation Datasets}
\label{supp:datasets}

We evaluate our CoMA on four widely-used benchmarks for domain adaptation, 
covering diverse scenarios from small-scale to large-scale transfers. Table~\ref{tab:dataset} summarizes the statistics of each dataset.

\vspace{4pt}
\noindent\textbf{Office-31}~\cite{office31} is a small-scale benchmark containing 
4,110 images across 31 object categories collected from three domains: 
Amazon (A), Webcam (W), and DSLR (D). We evaluate on all 6 transfer tasks.

\vspace{4pt}
\noindent\textbf{Office-Home}~\cite{officehome} is a medium-scale benchmark. It contains 
15,588 images spanning 65 categories across four distinct domains: 
Art (Ar), Clip Art (Cl), Product (Pr), and Real-World (Rw). This dataset presents greater domain diversity and visual complexity compared to Office-31. We conduct experiments on all 12 transfer 
tasks among the four domains.

\vspace{4pt}
\noindent\textbf{DomainNet-126}~\cite{domainnet-126} is a subset of the full DomainNet dataset, 
containing 126 categories across four domains: Clipart (C), Painting (P), 
Real (R), and Sketch (S), with a total of 142,334 images. 
This large-scale benchmark tests the scalability of adaptation methods to 
datasets with many classes. We evaluate on all 12 possible transfer tasks.

\vspace{4pt}
\noindent\textbf{VisDA-C}~\cite{visda-c} is a large-scale synthetic-to-real benchmark. It contains 12 object categories with 152,397 synthetic images rendered from 3D models (source domain) and 
55,388 real images (target domain). Following the standard protocol, we evaluate on the single Sy$\to$Re transfer task.

\begin{table}[t]
\centering
\caption{{\bf Statistics of evaluation datasets.} \#Image denotes the total number 
of images, \#Class denotes the number of categories, and \#Domain denotes 
the number of domains.}
\vspace{-0.15cm}
\label{tab:dataset}
\renewcommand\tabcolsep{1.8pt}
\renewcommand\arraystretch{0.9}
\scriptsize
\centering
\resizebox{\linewidth}{!}{
    \begin{tabular}{l|ccc|c}
    \toprule
    Dataset & \#Image & \#Class & \#Domain & Transfer tasks \\
    \midrule
    Office-31~\cite{office31} & 4.1k & 31 & 3 & 6 \\
    Office-Home~\cite{officehome} & 15.5k & 65 & 4 & 12 \\
    DomainNet-126~\cite{domainnet-126} & 142k & 126 & 4 & 12 \\
    VisDA-C~\cite{visda-c} & 207k & 12 & 2 & 1 \\
    \bottomrule
    \end{tabular}
    }
\end{table}

\section{Implementation Details}
\noindent\textbf{Source model pre-training.}
For all domain adaptation tasks across the four datasets, we pre-train the source model $\theta_s$ following the standard procedure of SHOT~\cite{shot}.  
Specifically, $\theta_s$ is trained in a supervised manner using cross-entropy loss on the labeled source data, formulated as:
\begin{equation*}
    \mathcal{L}(\mathcal{X}_s,\mathcal{Y}_s;\theta_s)
    = -\mathbb{E}_{(x_s, y_s) \in \mathcal{X}_s \times \mathcal{Y}_s}
    \sum_{k=1}^{K} \tilde{l}_{k}^s \log \theta_s(x_s),
\end{equation*}
where $\tilde{l}_{k}^s = (1-\sigma)l_{k}^s + \frac{\sigma}{K}$ denotes the label-smoothed~\cite{smooth} soft label, and $l_{k}^s$ represents the one-hot encoding of the ground-truth label $y_s \in \mathcal{Y}_s$.  
We set $\sigma = 0.1$ throughout all experiments.

\vspace{4pt}
\noindent\textbf{Burn-in stage.}
\label{supp:burn-in}
Due to BLIP~\cite{instblip}’s large model size and its caption-oriented prediction space, it is impractical to use the full model during adaptation.
Instead, we employ a proxy model $\theta_b$ trained on BLIP-generated pseudo-labels, serving as a functional substitute for BLIP during adaptation.

To train the BLIP-proxy model, we first generate pseudo-labels $\hat{y}_t^b$ from BLIP’s captions.
For each target image $x_t$, BLIP produces a caption $\Theta_b(x_t, v_b)$ with hand-crafted prompt $v_b$.
We then compute the cosine similarity between text embedding of the caption and the text embedding 
of each class name using a general text encoder $T(\cdot)$~\cite{gte}:
\begin{equation*}
\hat{y}_t^b = \underset{k}{\arg\max}
\frac{T(\Theta_b(x_t,v_b))^\top \cdot T(\mathrm{CLS}_k)}
{||T(\Theta_b(x_t, v_b))||\cdot||T(\mathrm{CLS}_k)||},
\end{equation*}
where $\mathrm{CLS}_k$ denotes the $k$-th class name.
Here, we use the following handcrafted prompt empirically: ``Describe the objects in this {\{domain\}}-image in the context of the following classes: $[\mathrm{CLS}_1,\mathrm{CLS}_2,\ldots,\mathrm{CLS}_K]$", where \{domain\} is replaced by the actual domain name (e.g., ``Clipart").
The proxy model $\theta_b$ adopts the same architecture as the source model $\theta_s$ and is initialized as $\theta_b = \theta_s$.
It is then trained on the generated pseudo-labels $\hat{y}_t^b$ via:
\begin{equation*}
\mathcal{L}(\mathcal{X}_t,\hat{\mathcal{Y}}_t^b;\theta_b)
= -\mathbb{E}_{(x_t,\hat{y}_t^b)\in(\mathcal{X}_t,\hat{\mathcal{Y}}_t^b)}
\sum_{k=1}^{K}\tilde{l}_{k}^b\log \theta_b(x_t),
\end{equation*}
where $\tilde{l}_{k}^b=(1-\sigma)l_{k}^b+\frac{\sigma}{K}$ denotes the label-smoothed soft label and $l_{k}^b$ is the one-hot encoding of $\hat{y}_t^b\in\hat{\mathcal{Y}}_t^b$.
We set $\sigma=0.1$ throughout our experiments.

\vspace{4pt}
\noindent\textbf{Network architecture.}
Our CoMA consists of three main components: a target model $\theta_t$, CLIP $\Theta_c$, and a 
BLIP-proxy model $\theta_b$.
For the target model, we adopt the same architecture as prior studies~\cite{difo,prode}, following standard protocol~\cite{shot}. The feature extractor comprises a deep convolutional network pre-trained on ImageNet, followed by a fully-connected bottleneck layer with batch normalization. Specifically, we use ResNet-50~\cite{resnet} for Office-31, Office-Home, and DomainNet-126, and ResNet-101~\cite{resnet} for VisDA-C. The classifier is a fully-connected layer with weight normalization.
For CLIP, we adopt the ViT-B/32 backbone as in~\cite{difo,prode}, where the image encoder uses Vision Transformer and the text encoder follows the transformer architecture proposed in~\cite{clip}. We initialize the learnable prompt context $v$ with ``a photo of a [CLS]'' and optimize it during training via prompt learning~\cite{coop}.
The BLIP-proxy model $\theta_b$ shares the same architecture as the target model and is trained with pseudo-labels generated by BLIP~\cite{instblip} during the burn-in phase.

\vspace{4pt}
\noindent\textbf{Training configuration.}
We train CoMA using SGD optimizer with momentum 0.9 and weight decay $1 \times 10^{-3}$. Unlike conventional MI objectives requiring large batches for sufficient class coverage, our DMI formulation ensures stable optimization even with small batches. This allows us to adopt dataset-specific batch configurations. We empirically set (batch size, epochs) as (16, 50) for Office-31, (32, 50) for Office-Home, (64, 20) for DomainNet-126, and (32, 15) for VisDA-C. All experiments use PyTorch on a single NVIDIA RTX A5000 (24GB).

\vspace{4pt}
\noindent\textbf{Hyper-parameter configuration.}
We set balancing coefficients $(\alpha, \beta) = (1.0, 0.6)$ for Office-31 and $(1.0, 0.5)$ for other datasets, where $\alpha$ and $\beta$ control TCA in Eq. (8) and MDA in Eq. (11) respectively. The DMI scale parameter is fixed to $\lambda = 0.5$ across all datasets (see Section~\ref{supp:extended_dmi}).
\begin{figure}[t] 
    \setlength{\belowcaptionskip}{-5pt}
    \setlength{\abovecaptionskip}{-2pt}
    \begin{center}
     \includegraphics[width=0.95\linewidth]{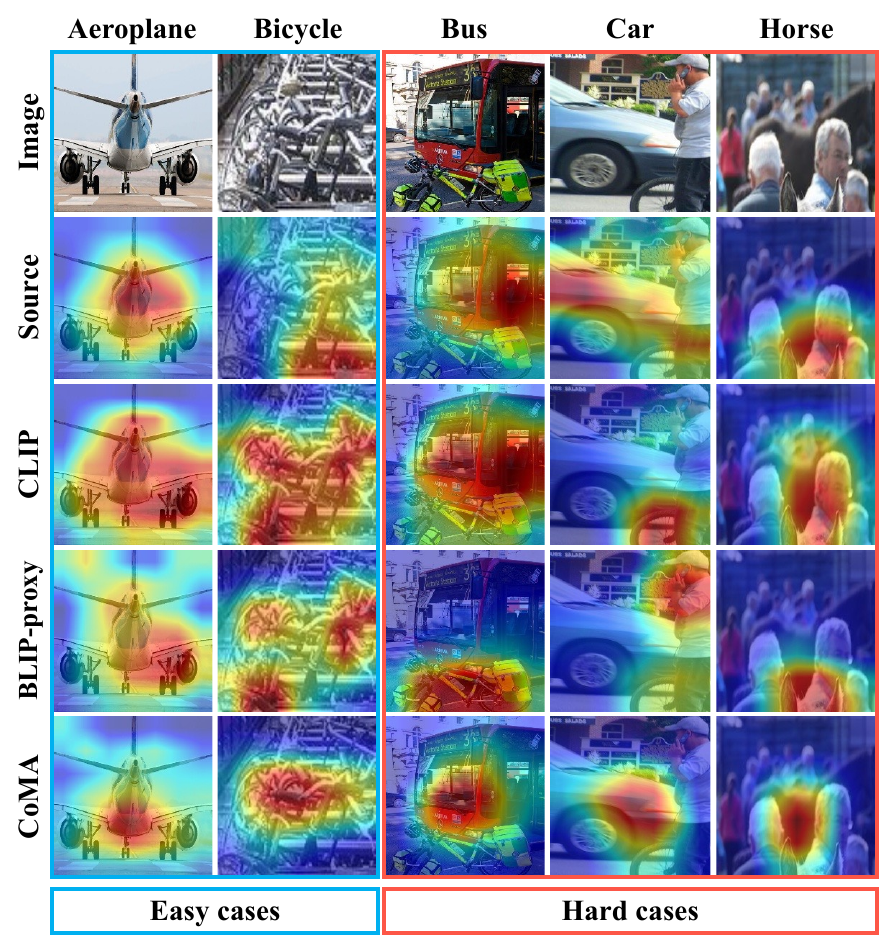}
    \end{center}
    \caption{
    Grad-CAM visualization of our CoMA and its three components on VisDA.
    }
    \vspace{-0.2cm}
    \label{fig:grad}
\end{figure}

\begin{table*}[t]
    \caption{{\bf Complete Closed-set SFDA accuracy} (\%) on {VisDA} across all 12 categories. \textbf{SF}, \textbf{C} and \textbf{B} indicate source-free, CLIP and BLIP, respectively. The best accuracy is indicated in \textbf{bold} and the second best one is \underline{underlined}.}
    \label{tab:VisDA}
    \vspace{-0.15cm}
    \renewcommand\tabcolsep{4.0pt}
    \renewcommand\arraystretch{0.9}
    \scriptsize
    \centering
    \resizebox{\textwidth}{!}{
        \begin{tabular}{ l l | c c c | c c c c c c c c c c c c | c}
        \toprule
        Method & Venue & \textbf{SF} & \textbf{C} & \textbf{B} & plane & bcycl & bus & car & horse & knife & mcycl & person & plant & sktbrd & train & truck & Avg.\\
        \midrule
        Source &-- &-- &-- &-- & 57.1 & 17.2 & 45.9 & 75.6 & 58.4 & 3.6 & 82.0 & 20.0 & 75.9 & 39.0 & 81.3 & 8.8 & 47.1 \\
        \midrule
        SHOT~\cite{shot} & ICML20 & \cmark & \xmark & \xmark & 94.3 & 88.5 & 80.1 & 57.3 & 93.1 & 94.9 & 80.7 & 80.3 & 91.5 & 89.1 & 86.3 & 58.2 & 82.9 \\
        NRC~\cite{nrc} & NIPS21 & \cmark & \xmark & \xmark & 96.8 & 91.3 & 82.4 & 62.4 & 96.2 & 95.9 & 86.1 & \underline{90.7} & 94.8 & 94.1 & 90.4 & 59.7 & 85.9 \\
        GKD~\cite{gkd} & IROS21 & \cmark & \xmark & \xmark & 95.3 & 87.6 & 81.7 & 58.1 & 93.9 & 94.0 & 80.0 & 80.0 & 91.2 & 91.0 & 86.9 & 56.1 & 83.0 \\
        AaD~\cite{aad} & NIPS22 & \cmark & \xmark & \xmark & 97.4 & 90.5 & 80.8 & 76.2 & 97.3 & 96.1 & 89.8 & 82.9 & 95.5 & 93.0 & 92.0 & 64.7 & 88.0 \\
        AdaCon~\cite{adacon} & CVPR22 & \cmark & \xmark & \xmark & 97.0 & 84.7 & 84.0 & 77.3 & 96.7 & 93.8 & 91.9 & 84.8 & 94.3 & 93.1 & 94.1 & 49.7 & 86.8 \\
        CoWA~\cite{cowa} & ICML22 & \cmark & \xmark & \xmark & 96.2 & 89.7 & 83.9 & 73.8 & 96.4 & 97.4 & 89.3 & 86.8 & 94.6 & 92.1 & 88.7 & 53.8 & 86.9 \\
        ELR~\cite{elr} & ICLR23 & \cmark & \xmark & \xmark & 97.1 & 89.7 & 82.7 & 62.0 & 96.2 & 97.0 & 87.6 & 81.2 & 93.7 & 94.1 & 90.2 & 58.6 & 85.8 \\
        PLUE~\cite{plue} & CVPR23 & \cmark & \xmark & \xmark & 94.4 & 91.7 & 89.0 & 70.5 & 96.6 & 94.9 & 92.2 & 88.8 & 92.9 & 95.3 & 91.4 & 61.6 & 88.3 \\
        CPD~\cite{cpd} & PR24 & \cmark & \xmark & \xmark & 96.7 & 88.5 & 79.6 & 69.0 & 95.9 & 96.3 & 87.3 & 83.3 & 94.4 & 92.9 & 87.0 & 58.7 & 85.5 \\
        TPDS~\cite{tpds} & IJCV24 & \cmark & \xmark & \xmark & 97.6 & 91.5 & 89.7 & 83.4 & 97.5 & 96.3 & 92.2 & 82.4 & \underline{96.0} & 94.1 & 90.9 & 40.4 & 87.6 \\
        \midrule
        DAPL-R~\cite{daplr} & TNNLS23 & \xmark & \cmark & \xmark & 97.8 & 83.1 & 88.8 & 77.9 & 97.4 & 91.5 & 94.2 & 79.7 & 88.6 & 89.3 & 92.5 & 62.0 & 86.9 \\
        PADCLIP-R~\cite{padclipr} & ICCV23 & \xmark & \cmark & \xmark & 96.7 & 88.8 & 87.0 & 82.8 & 97.1 & 93.0 & 91.3 & 83.0 & 95.5 & 91.8 & 91.5 & 63.0 & 88.5 \\
        ADCLIP-R~\cite{adclipr} & ICCVW23 & \xmark & \cmark & \xmark & 98.1 & 83.6 & \best{{91.2}} & 76.6 & 98.1 & 93.4 & \best{{96.0}} & 81.4 & 86.4 & 91.5 & 92.1 & 64.2 & 87.7 \\
        PDA-R~\cite{pdar} & AAAI24 & \xmark & \cmark & \xmark & 97.2 & 82.3 & 89.4 & 76.0 & 97.4 & 87.5 & \underline{95.8} & 79.6 & 87.2 & 89.0 & 93.3 & 62.1 & 86.4 \\
        DAMP-R~\cite{dampr} & CVPR24 & \xmark & \cmark & \xmark & 97.3 & 91.6 & 89.1 & 76.4 & 97.5 & 94.0 & 92.3 & 84.5 & 91.2 & 88.1 & 91.2 & 67.0 & 88.4 \\
        \midrule
        DIFO-V~\cite{difo} & CVPR24 & \cmark & \cmark & \xmark & 97.5 & 89.0 & \underline{90.8} & \underline{83.5} & 97.8 & 97.3 & 93.2 & 83.5 & 95.2 & 96.8 & 93.7 & 65.9 & 90.3 \\
        ProDe-V~\cite{prode} & ICLR25 & \cmark & \cmark & \xmark & \underline{98.3} & \underline{92.4} & 86.6 & 80.5 & \underline{98.1} & \underline{98.0} & 92.3 & 84.3 & 94.7 & \underline{97.0} & \underline{94.1} & \best{{75.6}} & \underline{91.0} \\
        \midrule
        \rowcolor{bestrow}
        \textbf{CoMA} & -- & \cmark & \cmark & \cmark & \best{{98.5}} & \best{{92.9}} & 89.2 & \best{{85.1}} & \best{{98.5}} & \best{{98.5}} & 95.4 & \best{91.5} & \best{{95.8}} & \best{{97.8}} & \best{{94.8}} & \underline{71.7} & \best{{92.5}} \\
        \bottomrule
        \end{tabular}
    }
\end{table*}

\section{Supplementary Analysis}
\noindent\textbf{Grad-CAM visualization.}
Fig.~\ref{fig:grad} shows Grad-CAM visualizations~\cite{gradcam} comparing attention patterns of the source model, CLIP, BLIP-proxy, and our CoMA on VisDA. We analyze two scenarios to demonstrate the complementary nature of dual MFMs and their harmonization in CoMA.

For easy cases with single-category objects (Aeroplane, Bicycle), CLIP and BLIP-proxy exhibit different semantic granularities: CLIP focuses on global, category-level semantics with broad spatial coverage, whereas BLIP-proxy captures local contextual semantics, attending to specific object parts with fine-grained detail. Through interaction among the three models, CoMA produces adjusted attention patterns that refine the focus regions, effectively balancing global and local perspectives.

For hard cases, where multiple object categories coexist with occlusion (Bus, Car, Horse), CLIP and BLIP-proxy attend to different objects due to their distinct semantic properties. CoMA achieves semantic adaptation, producing refined attention that selectively concentrates on the most discriminative regions. This demonstrates that CoMA effectively harmonizes the complementary semantic cues from both MFMs, thereby achieving more balanced and discriminative attention across diverse visual contexts.

\begin{table}[H]
\caption{{\bf Computational cost analysis} on Office-Home.}
\label{tab:computational_cost}
\vspace{-0.15cm}
\renewcommand\tabcolsep{4.0pt}
\renewcommand\arraystretch{0.9}
\scriptsize
\centering
\resizebox{\linewidth}{!}{
    \begin{tabular}{l|cc|c|c}
    \toprule
    \multirow{2}{*}{Method} 
    & \multicolumn{2}{c}{\bf Train}\vline 
    & \multicolumn{1}{c}{\bf Inference}\vline 
    & \multicolumn{1}{c}{\bf Accuracy}\\
    & time/iter (ms) & GPU Memory (GB) & time/iter (ms) & Avg. (\%)\\
    \midrule
    ProDe-V~\cite{prode} & 197 & 6.1 & 52 & 84.5\\
    CoMA & 327 & 9.1 & 52 & 90.7\\
    \bottomrule
    \end{tabular}
}
\end{table}

\vspace{4pt}
\noindent\textbf{Computational cost analysis.}
We analyze the computational cost of our CoMA compared to 
ProDe-V~\cite{prode} on Office-Home. Table~\ref{tab:computational_cost} 
reports the training time per iteration, training GPU memory consumption, inference time per iteration, and final accuracy, all measured on a single NVIDIA RTX A5000 (24GB) with Intel Xeon E5-2650 v4 CPU and 128GB RAM under identical hardware and software configurations. Compared to ProDe-V~\cite{prode}, CoMA requires 66\% more training time per iteration and 50\% additional memory, primarily due to dual MFM forward passes and optimizations. However, this additional cost is justified by two critical advantages: (1) no inference overhead, as only the target model is deployed, and (2) substantially higher accuracy (+6.2\%), demonstrating a favorable trade-off for practical applications.

\begin{figure}[t] 
    \setlength{\belowcaptionskip}{-5pt}
    \setlength{\abovecaptionskip}{-2pt}
    \begin{center}
     \includegraphics[width=0.90\linewidth]{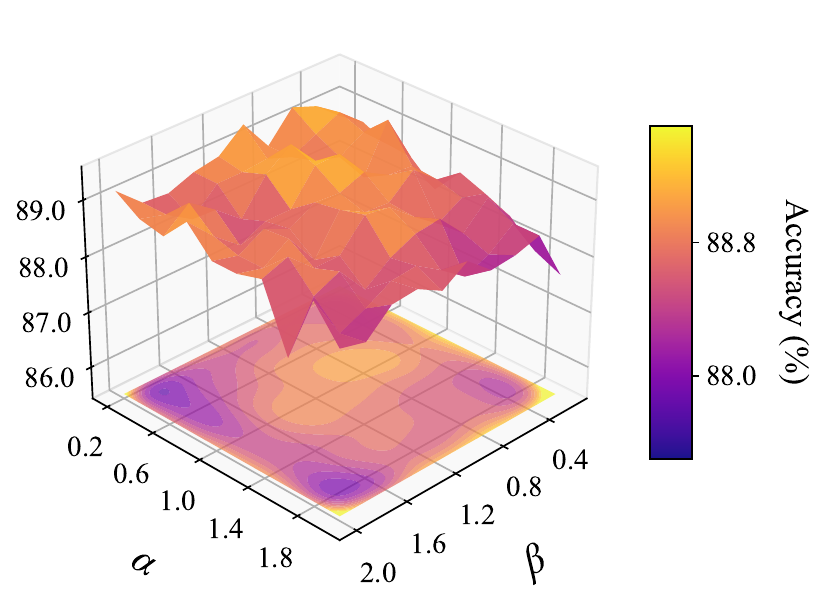}
    \end{center}
    \caption{
    Hyper-parameter sensitivity analysis of $\alpha,~\beta$ on the transfer task Cl$\to$Ar in Office-Home.
    }
    \label{fig:hyperparam}
\end{figure}

\vspace{4pt}
\noindent\textbf{Hyper-parameter sensitivity analysis.}
We analyze how the balancing coefficients $\alpha$ and $\beta$ affect our CoMA. 
As shown in Fig.~\ref{fig:hyperparam}, we vary both parameters over a broad range, $[0.2, 2.0]$, with a step size of 0.2.
Overall, the performance of CoMA varies only within a narrow band, demonstrating robustness to hyper-parameter choice. Noticeable degradation appears only at extreme values (either too small or too large), while most intermediate settings yield consistently stable accuracy. In particular, CoMA maintains strong performance when $\alpha$ lies within approximately $[0.6, 1.2]$ and $\beta$ within $[0.4, 1.0]$, confirming that our framework does not rely on precise hyper-parameter tuning.

\begin{table*}[t!]
    \caption{{\bf Complete Partial-set SFDA accuracy} (\%) on {Office-Home} across all 12 transfer tasks. The best accuracy is indicated in \textbf{bold} and the second best one is \underline{underlined}.}
    \label{tab:ps}
    \vspace{-0.15cm}
    \renewcommand\tabcolsep{2.5pt}
    \renewcommand\arraystretch{0.9}
    \scriptsize
    \centering
    \resizebox{\textwidth}{!}{
        \begin{tabular}{ll|cccc cccc cccc|c}
        \toprule
        Method &Venue &Ar$\to$Cl &Ar$\to$Pr &Ar$\to$Rw &Cl$\to$Ar &Cl$\to$Pr &Cl$\to$Rw &Pr$\to$Ar &Pr$\to$Cl &Pr$\to$Rw &Rw$\to$Ar &Rw$\to$Cl &Rw$\to$Pr &Avg. \\
        \midrule
        Source &-- &45.2 &70.4 &81.0 &56.2 &60.8 &66.2 &60.9 &40.1 &76.2 &70.8 &48.5 &77.3 &62.8 \\
        \midrule
        SHOT~\cite{shot} &ICML20 &64.8 &85.2 &92.7 &76.3 &77.6 &88.8 &79.7 &64.3 &89.5 &80.6 &66.4 &85.8 &79.3 \\
        HCL~\cite{hcl} &NIPS21 &65.6 &85.2 &92.7 &77.3 &76.2 &87.2 &78.2 &66.0 &89.1 &81.5 &68.4 &87.3 &79.6 \\ 
        AaD~\cite{aad} & NIPS22 &67.0 &83.5 &\underline{93.1} &80.5 &76.0 &87.6 &78.1 &65.6 &90.2 &83.5 &64.3 &87.3 &79.7 \\
        CoWA~\cite{cowa} &ICML22 &69.6 &\underline{93.2} &92.3 &78.9 &81.3 &\underline{92.1} &79.8 &\underline{71.7} &90.0 &83.8 &\underline{72.}2 &\underline{93.7} &83.2 \\
        CRS~\cite{crs} &CVPR23 &68.6 &85.1 &90.9 &80.1 &79.4 &86.3 &79.2 &66.1 &90.5 &82.2 &69.5 &89.3 &80.6 \\
        \midrule
        DIFO-V~\cite{difo} &CVPR24 &69.9 &88.8 &90.3 &\underline{85.7} &89.5 &91.2 &\underline{85.8} &70.3 &92.8 &87.1 &69.1 &89.1 & 84.1 \\
        ProDe-V~\cite{prode} &ICLR25 &\underline{70.2} &89.7 &90.4 &84.1 &\underline{90.7} &91.4 &85.5 &69.9 &\underline{92.9} &\underline{87.8} &68.5 &89.7 &\underline{84.2} \\
        \midrule
        \rowcolor{bestrow}
        \textbf{CoMA} & -- &\best{83.0} &\best{95.7} &\best{94.0} &\best{94.5} &\best{94.8} &\best{93.3} &\best{92.8} &\best{84.5} &\best{93.9} &\best{93.3} &\best{84.0} &\best{94.7} &\best{91.5}\\
        \bottomrule
        \end{tabular}
    }
\end{table*}

\begin{table*}[t!]
    \caption{{\bf Complete Open-set SFDA accuracy} (\%) on {Office-Home} across all 12 transfer tasks. The best accuracy is indicated in \textbf{bold} and the second best one is \underline{underlined}.}
    \label{tab:os}
    \vspace{-0.15cm}
    \renewcommand\tabcolsep{2.5pt}
    \renewcommand\arraystretch{0.9}
    \scriptsize
    \centering
    \resizebox{\textwidth}{!}{
        \begin{tabular}{ll|cccc cccc cccc|c}
        \toprule
        Method &Venue &Ar$\to$Cl &Ar$\to$Pr &Ar$\to$Rw &Cl$\to$Ar &Cl$\to$Pr &Cl$\to$Rw &Pr$\to$Ar &Pr$\to$Cl &Pr$\to$Rw &Rw$\to$Ar &Rw$\to$Cl &Rw$\to$Pr &Avg. \\
        \midrule
        Source &-- &36.3 &54.8 &69.1 &33.8 &44.4 &49.2 &36.8 &29.2 &56.8 &51.4 &35.1 &62.3 &46.6 \\
        \midrule
        SHOT~\cite{shot} &ICML20 &64.5 &80.4 &84.7 &63.1 &75.4 &81.2 &65.3 &59.3 &83.3 &69.6 &64.6 &82.3 &72.8 \\
        HCL~\cite{hcl} &NIPS21 &64.0 &78.6 &82.4 &64.5 &73.1 &80.1 &64.8 &59.8 &75.3 &78.1 &69.3 &81.5 &72.6 \\ 
        AaD~\cite{aad} & NIPS22 &63.7 &77.3 &80.4 &66.0 &72.6 &77.6 &69.1 &62.5 &79.8 &71.8 &62.3 &78.6 &71.8 \\
        CoWA~\cite{cowa} &ICML22 &63.3 &79.2 &85.4 &67.6 &83.6 &82.0 &66.9 &56.9 &81.1 &68.5 &57.9 &85.9 &73.2 \\
        CRS~\cite{crs} &CVPR23 &65.2 &76.6 &80.2 &66.2 &75.3 &77.8 &70.4 &61.8 &79.3 &71.1 &61.1 &78.3 &73.2 \\
        \midrule
        DIFO-V~\cite{difo} &CVPR24 &64.5 &\underline{86.2} &\underline{87.9} &68.2 &79.3 &86.1 &67.2 &62.1 &\underline{88.3} &71.9 &65.3 &84.4 &75.9\\
        ProDe-V~\cite{prode} &ICLR25 &\underline{75.9} &85.6 &\underline{87.9} &\best{81.3} &\underline{86.8} &\underline{87.2} &\best{81.1} &\underline{74.3} &86.3 &\best{83.0} &\underline{75.7} &\underline{86.1} &\underline{82.6}\\
        \midrule
        \rowcolor{bestrow}
        \textbf{CoMA} & -- &\best{79.9} &\best{93.7} &\best{89.6} &\underline{79.7} &\best{93.5} &\best{88.8} &\underline{80.2} &\best{77.6} &\best{90.0} &\underline{81.4} &\best{76.4} &\best{93.4} &\best{85.4} \\
        \bottomrule
        \end{tabular}
    }
\end{table*}

\begin{table*}[t!]
    \caption{{\bf Component-wise Closed-set SFDA performance summary} (\%) on Office-31, Office-Home, DomainNet-126, and VisDA. The best accuracy is indicated in \textbf{bold}.}
    \label{tab:end-to-end}
    \vspace{-0.15cm}
    \renewcommand\tabcolsep{4.5pt}
    \renewcommand\arraystretch{0.9}
    \scriptsize
    \centering
    \resizebox{\textwidth}{!}{
        \begin{tabular}{ll|ccc|c|cccc|c|cccc|c|c}
        \toprule
        \multirow{2}{*}{Method} 
        & \multirow{2}{*}{Venue} 
        & \multicolumn{4}{c}{\bf Office-31}\vline 
        & \multicolumn{5}{c}{\bf Office-Home}\vline 
        & \multicolumn{5}{c}{\bf DomainNet-126}\vline
        & \multicolumn{1}{c}{\bf VisDA}\\
        & & $\to$A & $\to$D & $\to$W & Avg. & $\to$Ar & $\to$Cl & $\to$Pr & $\to$Rw & Avg. & $\to$C & $\to$P & $\to$R & $\to$S &Avg. & Sy$\to$Re\\
        \midrule
        Source &-- & 62.1 &88.5 &84.1 &78.2 &56.7 &43.2 &68.3 &70.1 &59.6 &56.2 &53.5 &65.1 &48.3 &55.8 &47.1\\
        CLIP-V32~\cite{clip} & ICML21 & 77.8 & 82.1 & 80.5 & 80.1 & 77.8 & 63.6 & 87.1 & 87.8 & 79.1 & 78.7 & 77.7 & 90.2 & 74.7 & 80.3 & 87.1\\
        InstBLIP-XL~\cite{instblip} & NIPS24 & 84.8 & 83.7 & 87.2 & 85.2 & 81.0 &79.5 &92.1 &88.1 &85.2 &84.8 &82.6 &87.3 &79.6 &83.6 & 86.8\\
        \midrule
        \rowcolor{bestrow}
        \textbf{CoMA} & -- & \best{86.7} &\best{95.8} &\best{96.0} &\best{92.8} &\best{89.2} &\best{84.1} &\best{95.8} &\best{93.8} &\best{90.7} &\best{89.1} &\best{85.3} &\best{90.6} &\best{85.2} &\best{87.6} &\best{92.5}\\
        \bottomrule
        \end{tabular}
    }
\end{table*}

\section{Supplementary Comparison Result}
\noindent\textbf{Complete result on Closed-set SFDA on VisDA.}
In the main paper, only the average accuracy was reported on VisDA. Here we present detailed results across all 12 categories. 
As shown in Table~\ref{tab:VisDA}, our CoMA achieves consistently strong performance across classes and obtains the highest average accuracy among all baseline methods. 
In detail, CoMA attains the highest accuracy in 9 out of 12 categories, demonstrating that the improvements are not confined to a few favorable classes but are broadly observed across diverse categories.

\vspace{4pt}
\noindent\textbf{Complete results on Partial-set and Open-set SFDA.}
Tables~\ref{tab:ps}--\ref{tab:os} provide the complete accuracy for the Partial-set and Open-set SFDA settings on Office-Home, covering all 12 transfer tasks.
CoMA consistently achieves the highest average accuracy in both settings.
In the Partial-set scenario, CoMA obtains the best performance on all 12 tasks, while in the Open-set scenario, it delivers superior results on 9 out of 12 tasks. Notably, the strong Partial-set 
performance is attributed to our DMI formulation, which focuses on represented classes, effectively ensuring more stable adaptation under incomplete label spaces.

\noindent\textbf{Component-wise analysis on Closed-set SFDA.}
Table~\ref{tab:end-to-end} summarizes the component-wise performance of CoMA across four SFDA benchmarks. This analysis highlights how the individual components, including the source model, CLIP-V32~\cite{clip}, and InstBLIP-XL~\cite{instblip}, synergistically combine within our CoMA to yield improved adaptation performance. Compared to each component model on its own, CoMA consistently produces higher accuracy across all datasets. For instance, on Office-Home, CoMA achieves 90.7\% accuracy, substantially outperforming the source model (59.6\%), CLIP-V32 (79.1\%), and InstBLIP-XL (85.2\%), validating the effectiveness of our collaborative multi-foundation approach. Furthermore, the improvements appear across all transfer directions, indicating that CoMA provides reliable performance gains without depending on specific domains. This consistent behavior supports the robustness and practicality of our CoMA in diverse SFDA scenarios.